\documentclass[review,authoryear,11pt]{elsarticle}

\usepackage{amsfonts}
\usepackage{amssymb}
\usepackage{amsbsy} 
\usepackage{amsmath}
\usepackage{subfigure}
\usepackage{color}
\usepackage{array}
\usepackage{pifont}

\newcommand{\x}{{\mathbf x}}

\newcommand{\IG}{\includegraphics}

\newcommand{\red}[1]{\textcolor[rgb]{0,0,0}{#1}}

\usepackage{hyperref}
\hypersetup{
    bookmarks=true,         
    unicode=false,          
    pdftoolbar=true,        
    pdfmenubar=true,        
    pdffitwindow=false,     
    pdfstartview={FitH},    
    pdftitle={},    
    pdfauthor={Verrelst},     
    pdfsubject={GPR-BAT},   
    pdfcreator={Verrelst},   
    pdfproducer={Verrelst}, 
    pdfkeywords={keyword1} {key2} {key3}, 
    pdfnewwindow=true,      
    colorlinks=true,       
    linkcolor=blue,          
    citecolor=blue,        
    filecolor=blue,      
    urlcolor=blue           
}

\usepackage{multicol}
\usepackage{float}
\usepackage{lscape}
\usepackage{cite}
	\usepackage{tikz}
\usetikzlibrary{arrows}

\usepackage{algorithm}
\usepackage{algpseudocode}

\journal{International Journal of Applied Earth Observation and Geoinformation}

\begin{document}
\begin{frontmatter}

\title{Spectral band selection for vegetation properties retrieval using Gaussian processes regression\corref{cor1}}

\author[uv]{Jochem Verrelst}
\ead{jochem.verrelst@uv.es}
\author[uv,bc]{Juan Pablo Rivera}
\author[ag]{Anatoly Gitelson}
\author[uv]{Jesus Delegido}
\author[uv]{Jos\'e Moreno}
\author[uv]{Gustau Camps-Valls}

\cortext[cor1]{\bf Preprint. Paper published in International Journal of Applied Earth Observation and Geoinformation
Volume 52, October 2016, Pages 554-567. DOI: j.jag.2016.07.016}
\address[uv]{Image Processing Laboratory (IPL), Parc Cient\'ific, Universitat de Val\`encia, 46980 Paterna, Val\`encia, Spain.}
\address[bc]{Departamento de Oceanograf\'ia F\'isica, CICESE, 22860 Ensenada, Mexico}
\address[ag]{Israel Institute of Technology, Technion, Haifa, Israel.} 
\fntext[phones]{Tel.: +34-96-354-40-67; Fax: +34-96-354-32-61.}

\begin{abstract}
With current and upcoming imaging spectrometers, automated band analysis techniques are needed to enable efficient identification of most informative bands to facilitate optimized processing of spectral data into estimates of biophysical variables. This paper introduces an automated spectral band analysis tool (BAT) based on Gaussian \red{processes} regression (GPR) for the spectral analysis of vegetation properties. The GPR-BAT procedure sequentially backwards removes the least contributing band in the regression model for a given variable until only one band is kept. 
GPR-BAT is implemented within the framework of the free ARTMO's MLRA (machine learning regression algorithms) toolbox, which is dedicated to the transforming of optical remote sensing images into biophysical products.
GPR-BAT allows (1) to identify the most informative bands in relating spectral data to a biophysical variable, and (2) to find the least number of bands that preserve optimized accurate predictions. To illustrate its utility, two hyperspectral datasets were analyzed for most informative bands: (1) a field  hyperspectral dataset (400-1100 nm at 2 nm resolution: 301 bands) with leaf chlorophyll content (LCC) and green leaf area index (gLAI) collected for maize and soybean (Nebraska, US); and (2) an airborne HyMap dataset (430-2490 nm: 125 bands) with LAI and canopy water content (CWC) collected for a variety of crops (Barrax, Spain). 
For each of these biophysical variables, optimized retrieval accuracies can be achieved with just 4 to 9 well-identified bands, and performance was largely improved over using all bands. A PROSAIL global sensitivity analysis was run to interpret the validity of these bands. Cross-validated $R^2_{CV}$ (NRMSE$_{CV}$) accuracies for optimized GPR models were 0.79 (12.9\%) for LCC, 0.94 (7.2\%) for gLAI, 0.95 (6.5\%) for LAI and 0.95 (7.2\%) for CWC.
This study concludes that a wise band selection of hyperspectral data is strictly required for optimal vegetation properties mapping. 
\end{abstract}


\end{frontmatter}
 

\section{Introduction}\label{sec:intro}
 
A new era of optical remote sensing science is emerging with forthcoming space-borne imaging spectrometer missions such as EnMAP \red{(Environmental Mapping and Analysis Program)} \citep{Guanter2015}, HyspIRI \red{(Hyperspectral Infrared Imager)}~\citep{Roberts12}, PRISMA \red{(PRecursore IperSpettrale della Missione Applicativa)}~\citep{Labate2009} and ESA's $8\textsuperscript{th}$ Earth Explorer FLEX \red{(Fluorescence Explorer)}~\citep{Kraft2012}. Having access to operationally acquired imaging spectroscopy data with hundreds of bands paves the path for a wide variety of monitoring applications, such as the quantification of structural and biochemical vegetation properties~\citep{schaepman09,Ustin2010,Homolova2013}. 

Facing such exciting new technological opportunity poses, however, an important methodological problem. Imaging spectroscopy data include highly correlated and noisy spectral bands, and frequently create statistical problems \red{(e.g., the Hughes effect)} due to small sample sizes compared to the large number of available, possibly redundant, spectral bands. These characteristics may lead to a violation of basic assumptions behind statistical models or may otherwise affect the model outcome. Models fitted with such multi-collinear data sets are prone to over-fitting, and transfer to other scenarios may thus be limited. Naturally, these issues affect the prediction accuracy as well as the interpretability of the regression (retrieval) models~\citep{Curran1989,Grossman1996}. It may therefore be desirable to reduce the spectral dimension, either through spectral dimensionality reduction techniques \citep[e.g.,][]{VanDerMaaten2007,Arenas2013} or to select particular spectral regions that are most helpful to describe targeted biophysical variables.
Apart from improving the fit and processing speed of regression models, selecting specific spectral regions may allow clarification of the relationships of spectral signatures to leaf and canopy optical properties while minimizing the signal from secondary responses~\citep{Feilhauer2015}.

From a pure statistical signal processing point of view, band selection is cast as an optimization problem by which one wants to select a subset of spectral wavebands that capture most of the information for a particular problem. The search for the best bands out of the available is known to be an NP-complete problem (it cannot be solved in polynomial time)~\citep{blum97} and the number of local minima can be quite large. This poses both numerical and computational difficulties.
 
Following a general taxonomy, {\em band selection} can be divided into two major categories: filter methods~\citep{liu98} and wrapper~\citep{kohavi97} methods. {\em Filter methods} use an indirect measure of the quality of the selected bands, so a faster convergence of the algorithm is obtained. {\em Wrapper methods} use the output of the learning machine as selection criteria. This approach guarantees that in each step of the algorithm, the selected subset improves the performance of the previous one. Filter methods might fail to select the right subset of bands if the used criterion deviates from the one used for training the learning machine, whereas wrapper methods can be computationally intensive since the learning machine has to be retrained for each new set of bands. 

Spectral band selection for quantifying vegetation properties have used both filter and wrapper band selection methods. Although there are many works of feature (spectral band) selection in imaging spectroscopy and remote sensing, however, the vast majority are related to classification problems \citep[e.g.,][]{Bazi2006,Archibald2007,Pal2010}; very few are concerned with regression (retrieval) problems, and in particular with vegetation properties estimation.

On the one hand, filter methods have long been restricted to the systematic calculation of all possible band combinations, e.g. through generic vegetation indices where all bands are combined into two-band indices and then applied to regression~\citep[e.g.,][]{Heiskanen13,Rivera2014}. However, these are brute-force techniques that usually do not go beyond searching for (linear or polynomial) combinations of two or at most three bands that maximize a fitting criterion (typically linear correlation). On the other hand, wrapper methods have been also applied in the field of chemometrics~\citep{Forina2004,Andersen2010}. 
 
Here, the focus is on non-parametric, multivariate regression methods. These are full-spectrum statistical methods, and some of them have band ranking properties through wrapper methods. There is a large evidence of the successful performance. For instance,~\citet{Feilhauer2015} compared three multivariate regression techniques (partial least square regression, random forests regression, and support vector regression) in their suitability for the identification and selection of spectral bands. A multi-method ensemble strategy, i.e. decision fusion, using these three methods was proposed in order to crystallize a more robust band selection. Among preferred univariate regression methods we find random forests~\citep{Genuer2010} mostly embedded in genetic algorithm procedures~\citep{Jung13}, or via permutation analyses.

A drawback of the above wrapper methods is that they are very often perceived as complex, e.g. they require software packages and method tuning is mostly needed, and not all of these methods performed equally well~\citep{Feilhauer2015}. Using a  regression method with few hyper-parameters to be tuned is perhaps the main problem here, and alternatives exist. Actually, various alternative non-parametric multivariate methods in the field of machine learning regression algorithms (MLRAs) equally possess band selection/ranking features, which some of them are very competitive. Comparison studies have demonstrated that the above-mentioned methods may not always be most powerful regression algorithms~\citep{Verrelst12a,Rivera13b, verrelst2015}. In these studies, it was shown that Gaussian processes regression (GPR)~\citep{Rasmussen06} outperformed other MLRAs for the retrieval of biophysical variables from airborne and satellite images~\citep{Verrelst12a,verrelst2015}. Of interest is that GPR also provides band ranking feature, which reveals the bands that contribute most to the development of a GPR model~\citep{CampsValls16grsm}. Given its powerful performance, GPR may be a first choice to exploit band ranking features.

Altogether, apart from above and a few more experimental studies \citep[e.g.,][]{Verrelst12a,Verrelst2012,Wittenberghe2014}, band ranking has not been fully exploited in retrieval applications. So far all these studies are experimental, and - while having their scientific merits - none of these methods are directly applicable to operational processing of hyperspectral data streams. For instance,  in view of optimized vegetation properties mapping, no user-friendly software package enabling automated identification of most important spectral bands for a given biophysical variable is available to the broader community. Such kinds of tools may become critical when forthcoming unprecedented hyperspectral data stream will become freely accessible. 

The objectives of this work are therefore threefold: (1) to develop a GPR-based band analysis tool, further referred to as ``GPR-BAT'', that analyzes the band-specific information content of spectral data for a given biophysical variable with little user interaction; (2) 
to demonstrate GPR-BAT's utility by applying it to two extensive hyperspectral datasets (biophysical variables and associated spectra) in order to identify the optimal number of bands and their spectral location; and finally, (3) to apply GPR-BAT to  hyperspectral image datasets for automated and optimized vegetation properties mapping. GPR-BAT will be operated as a graphical user interface (GUI) within  ARTMO's (automated radiative transfer models operator)~\citep{Verrelst12c} machine learning regression algorithm (MLRA) toolbox~\citep{Rivera13b}. To assess the optimality of the identified bands, we will run a global sensitivity analysis applied to the physically-based PROSAIL canopy radiative transfer model (RTM).

\section{Gaussian processes regression}\label{sec:GPR}

\section{ARTMO and GPR-BAT}\label{sec:ARTMO}
We here summarize the main characteristics of the ARTMO toolbox, and the GPR band analysis tool for automatic band selection implemented in ARTMO.

\subsection{ARTMO: a toolbox for automated vegetation properties retrieval}

This work contributes to the expansion of the in-house developed software package  ARTMO~\citep{Verrelst12c}. ARTMO embodies a suite of leaf and canopy radiative transfer models (RTMs) and several retrieval toolboxes, i.e. a spectral indices toolbox,~\citep{Rivera2014}, a LUT-based inversion toolbox~\citep{Rivera13a}, and a machine learning regression algorithm (MLRA) toolbox~\citep{Rivera13b}. 
The MLRA retrieval toolbox offers a suite of regression algorithms that enable to estimate biophysical parameters based on either experimental or simulated data in a semiautomatic fashion. The MLRA toolbox is essentially built around the SimpleR package~\citep{simpler} with over 15 non-parametric regression algorithms, including GPR, along with cross-validation training/validation sub-sampling and dimensionality reduction techniques. In this latest version (v1.17), the GPR-based band analysis tool (GPR-BAT) is for the first time introduced. The ARTMO package runs in MATLAB and can be downloaded at: ~\url{http://ipl.uv.es/artmo/}.

\subsection{GPR band analysis tool (GPR-BAT)}

\begin{figure}[t!]
\begin{center} 
\IG[width=12cm]{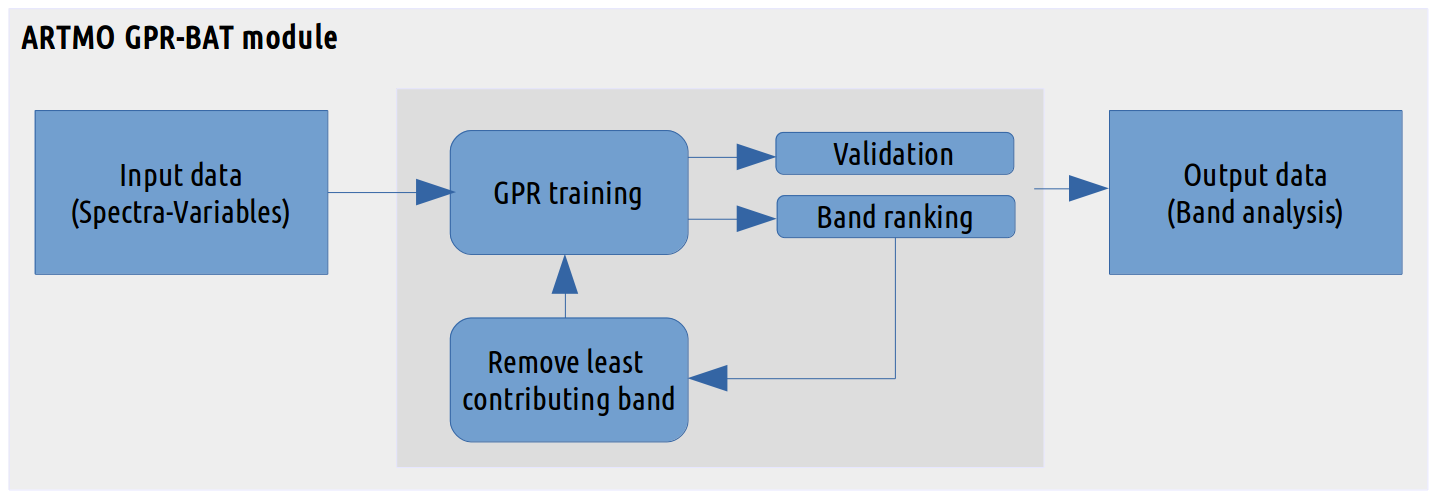} 
\end{center}
\caption{Schematic flow diagram of GPR-BAT within ARTMO's MLRA toolbox. }
\label{GPR_flowchart}
\end{figure}

As outlined above, one of the advantages of GPR is that during the development of the GPR model the predictive power of each single band is evaluated for the variable of interest through calculation of the $\sigma_b$. Specifically, band ranking through $\sigma_b$ may reveal the bands that contribute most to the development of a GPR model. Accordingly, when removing the least contributing band (i.e. highest $\sigma_b$) and then again training and validating a {\em new} GPR model, this procedure can be repeated until eventually only one band is left. Consequently, the SBBR routine eventually leads to identification of most sensitive final band for the variable under consideration, and so provides a set ideal band combinations for any number of bands.  
Once the SBBR routine start running, band rankings and goodness-of-fit statistical outputs (e.g., coefficient of determination: $R^2$; root mean square error: RMSE; normalized RMSE: NRMSE) are directly stored in a MySQL database (see also~\citet{Rivera13b}). Results can be subsequently queried and plotted through an output GUI. Additionally, to ensure robust identification of most sensitive bands as well validation results, the method can be combined with a $k$-fold cross-validation (CV) sub-sampling scheme. This scheme first splits randomly the training data into $k$ mutually exclusive subsets (folds) of equal size and then by training $k$ times a regression model with variable-spectra pairs. Each time, we left out one of the subsets from training and used it (the omitted subset) only to obtain an estimate of the regression accuracy ($R^2$, RMSE, NRMSE). From $k$ times of training and validation, the resulting validation accuracies were averaged and basic statistics calculated (standard deviation, min-max) to yield a more robust validation estimate of the considered regression model (see also~\citet{verrelst2015}).
The $\sigma_b$ band rankings are also stored during the $k$-fold repetitions. Two band removal options are provided to proceed with SBBR, either based on (1) sum of $\sigma_b$ values, or (2) sum of $\sigma_b$ rankings. The $k$-fold CV SBBR routine is summarized in Algorithm \ref{GPR-BAT_k}. A schematic flow diagram of the GBP-BAT tool within the MLRA toolbox is provided in Figure \ref{GPR_flowchart}.

\begin{algorithm}[h!]
\caption{Sequential backward band removal with $k$-fold cross-validation subsampling}
\label{GPR-BAT_k}
\begin{algorithmic}[1]
\For{{Number of bands $B$}}
\State Resample spectra-variable dataset in $k$-fold subsets 
\For{{Each $k$-fold iteration}}
\State Train GPR model with $k$-fold training subset
\State Rank $\sigma_b$ obtained during GPR model development
\State Calculate score indicators in validation: $R^2$, RMSE, NRMSE
\EndFor
\State Sum/Rank the $\sigma_b$ of the $k$-fold subset 
\State Calculate statistics (mean, SD, min-max) from the $k$-fold score
\State Remove band with highest $\sigma_b$ from dataset
\EndFor
\end{algorithmic}
\end{algorithm}

The main purpose of GPR-BAT is to identify how many bands are minimally needed in order to retain robust results and what are the most sensitive wavelengths. Accordingly, the output GUI delivers the following band analysis outputs: (1) goodness-of-fit validation statistics as a function of \#bands plotted over the sequentially removed bands until only 1 band is left, (2) associated wavelengths, (3) if $k$-fold CV sub-sampling is applied, then it also provides frequency plotting of most relative bands (e.g. top 5) for a given number of bands (e.g. all bands). Finally, note that although in this paper emphasis is on vegetation properties, essentially GPR-BAT can be  applied to any measured (or modeled) surface biophysical or geophysical variable when associated with spectral data. 

\section{Global Sensitivity Analysis}\label{sec:GSA_theory}
Bearing in mind that GPR-BAT is a statistical, data-driven method, it is understood that the meaningfulness of the obtained best performing bands are entirely dependent on the quality of the introduced dataset. Here, to verify the significance of best-performing bands, they will be compared against a variance-based global sensitivity analysis (GSA) of the popular canopy radiative transfer model PROSAIL (PROSPECT4 + SAIL)~\citep{Jacquemoud2009}. Variance-based GSA explores the full input variable space and evaluates the relative importance of each input variable in a model~\citep{saltelli1999}. The method can be used to identify the most influential variables affecting model outputs. In variance-based GSA, the contribution of each input variable to the variation in outputs is averaged over the variation of all input variables, i.e., all input variables are changed together~\citep{saltelli1999}. Recently a GSA toolbox has been developed within the ARTMO package, where the GSA calculation of RTMs embedded in ARTMO (e.g. PROSAIL) has been largely automated~\citep{Verrelst2015d}. The method of~\citet{saltelli2010} was implemented, which has been demonstrated to be effective in identifying both the main sensitivity effects (first-order effects, i.e., the contribution to the variance of the model output by each input variables, $S_{i}$) and total sensitivity effects (the first-order effects plus interactions with other input variables, $S_{Ti}$) of input variables~\citep[e.g.,][]{Verrelst2015a,Verrelst2016}. 

GSA was applied to PROSAIL with vegetation input variables ranging within the min-max boundaries as listed in Table \ref{PROSAIL_inputs}. Variables were sampled according to Latin Hypercube sampling~\citep{McKay1979}. In total, (N(k + 2)) model simulations were run, where N is the sample size and equals 1000, and k is the number of input variables and equals 7. This produced 9000 simulations with directional reflectance at outputs between 400 and 2500 nm at 1 nm increments. Only total order sensitivity effects ($S_{Ti}$) expressed as percentages were considered.

\begin{table}[t!] 
\begin{center}
    \caption{Boundaries of used input variables of the PROSAIL (PROSPECT-4 + SAIL) model. \red{Non-vegetation} SAIL variables were kept to their default values, with a solar zenith angle of 30$^{\circ}$.}
    \resizebox{0.80\textwidth}{!}{ 
     \begin{tabular}{llccc@{}}
\hline
       \multicolumn{2}{c}{\bf Model variales} & {\bf Units} & {\bf Minimum} & {\bf Maximum}\\
\hline			
\multicolumn{5}{l}{$Leaf$ $variables$: PROSPECT-4} \\
$N$ &  Leaf structure index & unitless & 1.2 & 2.6 \\
LCC & Leaf chlorophyll content & [${\mu}$g/cm$^{2}$]  & 0 & 80 \\
$C_m$ &  Leaf dry matter content & [g/cm$^{2}$] & 0.001 & 0.02 \\
LWC &  Leaf water content & [g/cm$^{2}$] & 0.001 & 0.05 \\
\\
\multicolumn{5}{l}{$Canopy$ $variables$: SAIL}\\
LAI & Leaf area index & [m$^{2}$/m$^{2}$] & 0 & 7 \\
 
LAD & Leaf angle distribution & [$^{\circ}$] & 30 & 60 \\ 
\hline	
\\
	   \end{tabular}}
	   \label{PROSAIL_inputs}
 
\end{center}
\end{table}

\section{Case studies}\label{sec:expsetup}
To illustrate the utility of GPR-BAT, two datasets are presented. First a field hyperspectral dataset collected in Nebraska, US, and second an airborne campaign over Barrax, Spain named as `SPARC'.

\subsection{UNL field hyperspectral dataset}
\subsubsection{Study site}

The study site was located at the University of Nebraska-Lincoln (UNL) Agricultural Research and Development Center near Mead, Nebraska (41$^\circ$10"46.8'N, 96$^\circ$26"22.7'W, 361 m above mean sea level), which is a part of the AmeriFlux network. This study site consists of three approximately 65-ha fields. Each field was managed differently as either continuous irrigated maize, irrigated maize/soybean rotation, or rain-fed maize/soybean rotation following the best management practices (e.g. fertilization, herbicide/pesticide treatment) for eastern Nebraska for its respective planting cycle. There were a total of 16 and 8 field-years for maize and soybean, respectively, which had maximal green leaf area index (gLAI) values ranging from 4.3 to 6.5 $m^2/m^2$ for maize and 3.0 to 5.5 $m^2/m^2$ for soybean. Specific details of these three fields (Ameriflux sites Ne-1, Ne-2 and Ne-3) can be found in~\citep{verma2005,vina2011}.

\subsubsection{Field measurements}
Although a whole set of vegetation variables were measured, two important varables are considered in this study: leaf chlorophyll content (LCC) and gLAI.
LCC was measured using Red Edge Chlorophyll Index, ($CI_{red \thinspace  edge}$), ~\citep{gitelson2003} that relates leaf reflectance in the red edge ($R_{red \thinspace edge}$) and near infra-red ($R_{NIR}$) wavebands with pigment content:
\begin{equation}
CI_{red \thinspace  edge} = (R_{NIR}/R_{red \thinspace edge})-1,
\end{equation}
where $R_{NIR}$ is average leaf reflectance in the range from 770 through 800 nm and $R_{red \thinspace edge}$ is the average reflectance in the range from 720 to 730 nm. 
	
During the growing season, maize and soybean leaves within a wide range of greenness were collected from the crop fields and their reflectance was measured in the spectral range from 400 to 900 nm using a leaf clip, with a 2.3-mm diameter bifurcated fiber-optic cable attached to both an Ocean Optics USB2000 spectroradiometer and to an Ocean Optics LS-1 tungsten halogen light source (details in~\citet{gitelson2006,ciganda2009}). The leaf clip allows individual leaves to be held with a 60$^\circ$ angle relative to the bifurcated fiber-optic. The software CDAP (CALMIT, University of Nebraska-Lincoln Data Management Program) was used to acquire and process the data from the sensor. A Spectralon reflectance standard (99\% reflectance) was scanned before each leaf measurement. The reflectance at each wavelength was calculated as the ratio of upwelling leaf radiance to the upwelling radiance of the standard. The average reflectance obtained from 10 scans was used to compute the $CI_{red \thinspace  edge}$. Once these measurements were completed, two to four circular disks (1-cm diameter) were punched from each leaf for analytical extraction of LCC and quantification using absorption spectroscopy.  The extraction of LCC was done using 10 mL of 80\% acetone. The extinction absorption coefficients published by~\citet{porra1989} were used for final calculations of total LCC. 
	
Relationship between  $CI_{red \thinspace  edge}$ and LCC was established:  
\begin{equation}
LCC (mg/m^2) = 37.9+1353.7*CI_{red \thinspace  edge},
\end{equation}
and validated~\citep{gitelson2006,ciganda2009}.  Predicted LCC in maize and soybean leaves was closely linearly related to LCC measured analytically with RMSE $<$ 38 $mg/m^2$. LCC and NRMSE below 4.5\%. Then, this equation was used for calculating LCC using reflectances measured in both crops in the red edge and NIR bands. 

Located in each field there were six 20 x 20 m plots that represented major soil and crop production zones ~\citep{verma2005}. For estimating gLAI, 6$\pm$2 plants were selected from one or two rows totaling 1 m length within each plot every 10-14 days. Sampling rows were alternated between collections to minimize edge effects. Samples were transported on ice prior to gLAI measurements using an area meter (Model LI-3100, LI-COR, Inc., Lincoln, NE). gLAI measurements were determined by multiplying the green leaf area per plant by the plant population in the plot. The field-level gLAI was determined from the average of the plot gLAI measurements on a given sampling date (details are in~\citet{vina2011,nguy2012}).

\subsubsection{Spectral measurements}

The hyperspectral data was collected from 2001 through 2008 using an all-terrain sensor platform ~\citep{rundquist2004,Rundquist2014}. Two USB2000 (Ocean Optics, Inc.) radiometers with a dual
fiber system were used to collect top-of-canopy (TOC) reflectance in spectral range from 400 to 1100 nm with 2.0 nm 
resolution (301 bands). The downwelling fiber was fitted with a cosine diffuser to measure irradiance while the upwelling fiber collected radiance. The upwelling fiber had a field of view of approximately 2.4 m in diameter since the height of the fiber was maintained at a constant height above the canopy throughout the growing season. The median of 36 reflectance measurements collected along access
 roads into each field was used as the field-level reflectance measurement. A total of 278 spectra for maize and 145 for soybean were acquired over the study period (details are in~\citet{vina2011,nguy2012}). 
 The reflectance measurements and field measurements were not always concurrent and, since LAI changes gradually, spine interpolations were taken between destructive LAI sampling dates for each field in each year using Matlab.

\subsection{SPARC field and HyMap dataset}

\subsubsection{Study site}
The second experiment involves an experimental dataset. The widely used SPARC dataset~\citep{Delegido2013} was chosen. 
The SPectra bARrax Campaign (SPARC) field dataset encompasses different crop types, growing phases, canopy geometries and soil conditions. The SPARC-2003 campaign took place from 12 to 14 July in Barrax, La Mancha, Spain (coordinates 30$^\circ$3'N, 28$^\circ$6'W, 700 m altitude). Bio-geophysical parameters have been measured within a total of 108 Elementary Sampling Units (ESUs) for different crop types (garlic, alfalfa, onion, sunflower, corn, potato, sugar beet, vineyard and wheat). An ESU refers to a plot, which is sized compatible with pixel dimensions of about 20~m $\times$ 20~m. In the analysis no differentiation between crops was made. 

\subsubsection{Field measurements} 
Although a whole set of vegetation variables were measured, two important \red{variables} are considered in this study: LAI and canopy water content (CWC).
LAI has been derived from canopy measurements made with a LiCor LAI-2000 digital analyzer. Each ESU was assigned one LAI value, obtained as a statistical mean of 24 measurements (8 data readings $\times$ 3 replica) with standard errors ranging from 5\% to 10\%. 
Strictly speaking, assuming  a random leaf angle distribution, the impact of clumping has been assessed only partially using the LiCor and its corresponding software.  Hence, effective LAI is given as an output variable. For all ESUs, LAI ranges from 0.4 to 6.2 (m$^2$ single sided leaf surface)/(m$^2$ ground surface). 

CWC is the product of LAI with leaf canopy content (LWC). LWC is measured as follows. First dry and fresh matter content is weighted for a number of leaves (3 per ESU). From the two masses and the known sampled area, wet and dry biomass can be calculated. LWC is then calculated as the mass of water in a plant sample divided by the mass of the entire plant sample before drying (i.e. on a wet biomass basis). Units of LWC are in g/m$^2$ single sided leaf surface, while CWC is expressed as g/m$^2$ ground surface.

\subsubsection{Spectral measurements}
During the campaign, airborne hyperspectral HyMap flight-lines were acquired for the study site, during the month of July 2003. HyMap flew with a configuration of 125 contiguous spectral bands, spectrally positioned between 430 and 2490 nm. Spectral bandwidth varied between 11 and 21 nm. The pixel size at overpass was 5 m. The flight-lines were corrected for radiometric and atmospheric effects according to the procedures of~\citet{Guanter05}. Finally, a TOC reflectance dataset was prepared, referring to the centre point of each ESU and its corresponding LAI values.

\subsection{Experimental setup}


To ensure robust identification of sensitive bands related to vegetation properties, for each dataset a $k$-fold CV SBBR procedure was applied. That means that each iteration was $k$ times repeated but with different pools of training/validation dataset in such way that all samples are used for validation. In order that sufficient data remains into the validation subsets, the $k$ was set to 10 for the Nebraska dataset (a total of 263 samples) and to 4 for the SPARC dataset (a total of 100 samples), i.e. in both cases leading to validation subsets of about 25 samples. 

Goodness-of-fit validation statistics are then averaged for the $k$ validation subsets, i.e., $R^2_{CV}$, RMSE$_{CV}$, NRMSE$_{CV}$, and also associated standard deviation (SD) and min-max rankings are stored into MySQL tables. Based on these $k$ repetitions, the generated $\sigma_b$ were $k$ times ranked. When adding up these $k$ rankings for each band, the least contributing band was then removed, and so the analysis iteratively proceeded until eventually only one band remained (Algorithm \ref{GPR-BAT_k}). Processing speed of the training and validation of the GPR models were logged as well. All analysis was performed using a 64 bit processor (Intel Core$\texttrademark$ i7-4700MQ CPU@ 3.60 GHz, 16 GB RAM).


\section{Experimental results}\label{sec:results}
This section is devoted to analyze the band selection subsets as obtained by GPR-BAT in two particular settings. First we study the band rankings in the UNL field hyperspectral dataset, and then in the SPARC airborne hyperspectral dataset, leading to optimized maps of two vegetation properties. Finally, sensitive bands for crystallized best models are interpreted against PROSAIL GSA results.

\subsection{UNL field hyperspectral dataset: LCC and gLAI retrieval} 

\begin{figure}[h!]
\begin{center}
\begin{tabular}{cc}
LCC & gLAI \\
\IG[width=6.0cm, trim={0.3cm 0.1cm 0.3cm 0.3cm}, clip]{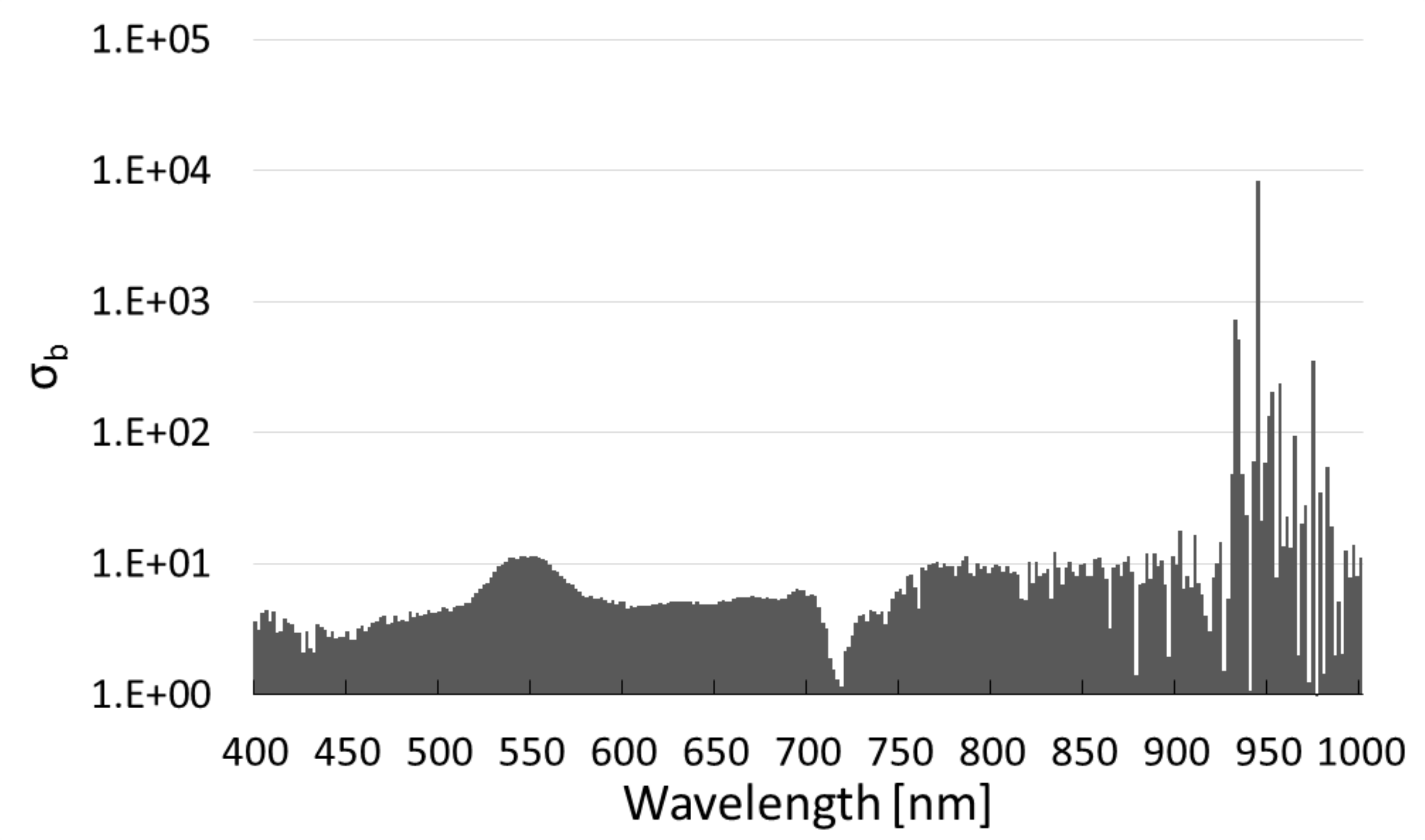} & 
\IG[width=6.0cm, trim={0.3cm 0.1cm 0.3cm 0.3cm}, clip]{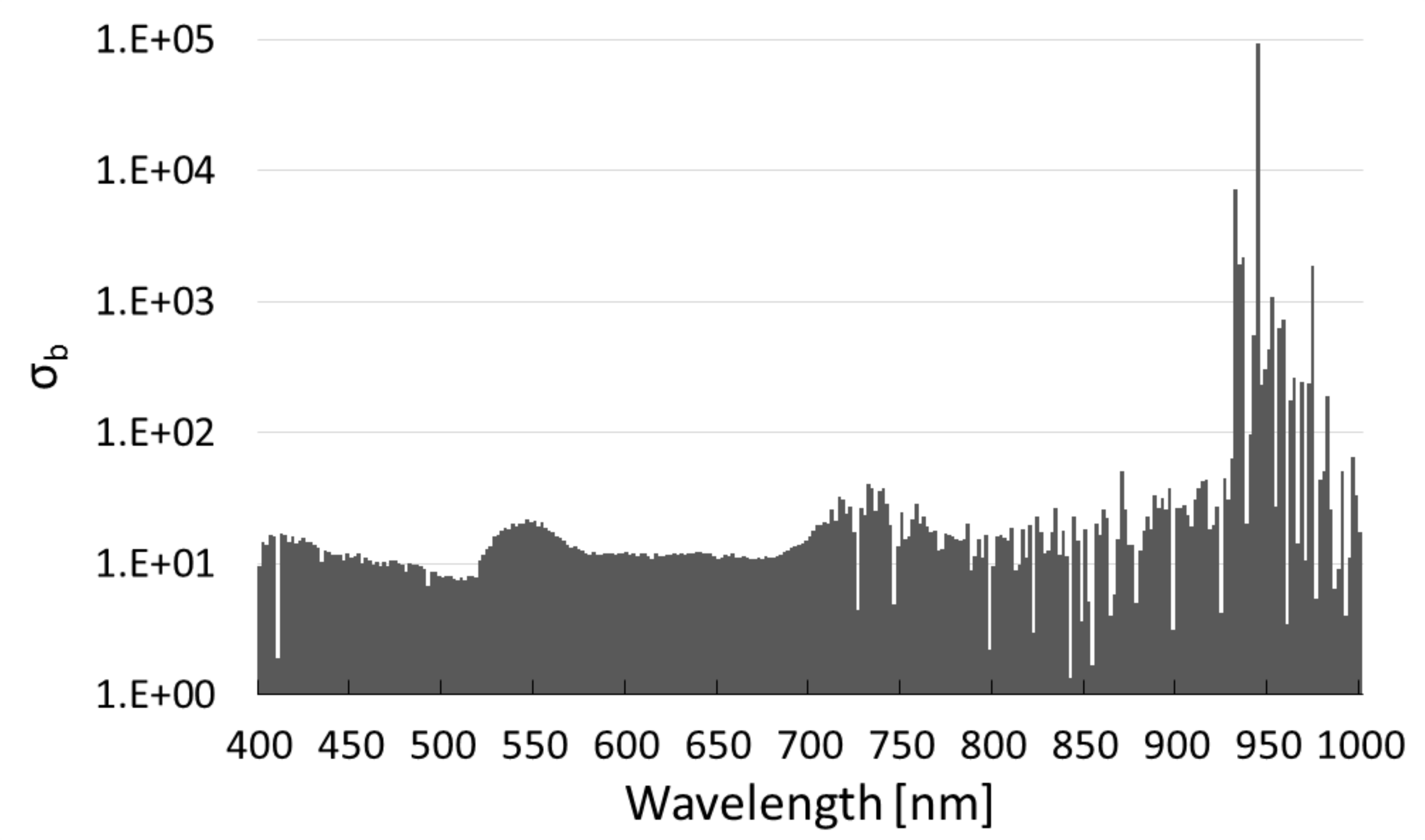}  \\
\end{tabular}
\end{center}
\caption{\red{Obtained $\sigma_b$ band rankings} of all USB2000 301 bands for single GPR model for LCC (left) and gLAI (right) plotted in logarithmic scale. The lower the $\sigma_b$ the more informative the band.}
\label{sigmas_all}
\end{figure}

The original full-spectrum radiometric data at 2 nm resolution was firstly analyzed. To be acquainted with GPR $\sigma_b$ band ranking, first the obtained $\sigma_b$ values for a single 10-fold GPR model for LCC and gLAI trained with all hyperspectral bands are illustrated in Figure \ref{sigmas_all}. Large difference in the $\sigma_b$ values can be observed for both variables. Noteworthy is that there are few spectral regions with informative bands (low $\sigma_b$) and particularly poorly informative bands (high $\sigma_b$). A distinct informative region is remarkable for LCC, where a systematic region of low $\sigma_b$ in the red edge (around 720 nm) appeared. At the same time, the $\sigma_b$ of poorly informative bands can rise to very high values (here plotted in logarithmic scale), which suggests that these bands can be problematic in the regression algorithm. However, since GPR is a data-driven algorithm, and the $\sigma_b$ depends on what has been presented during training phase, therefore, they may fluctuate depending on the given training-validation partitioning. It supports the rationale of using the SBBR procedure in combination with a $k$-fold CV sampling scheme to infer best performing bands. 

\begin{figure}[h!]
\begin{center}
\begin{tabular}{cc}
LCC & gLAI \\
\IG[width=6.0cm, trim={1cm 1cm 4cm 1cm}, clip]{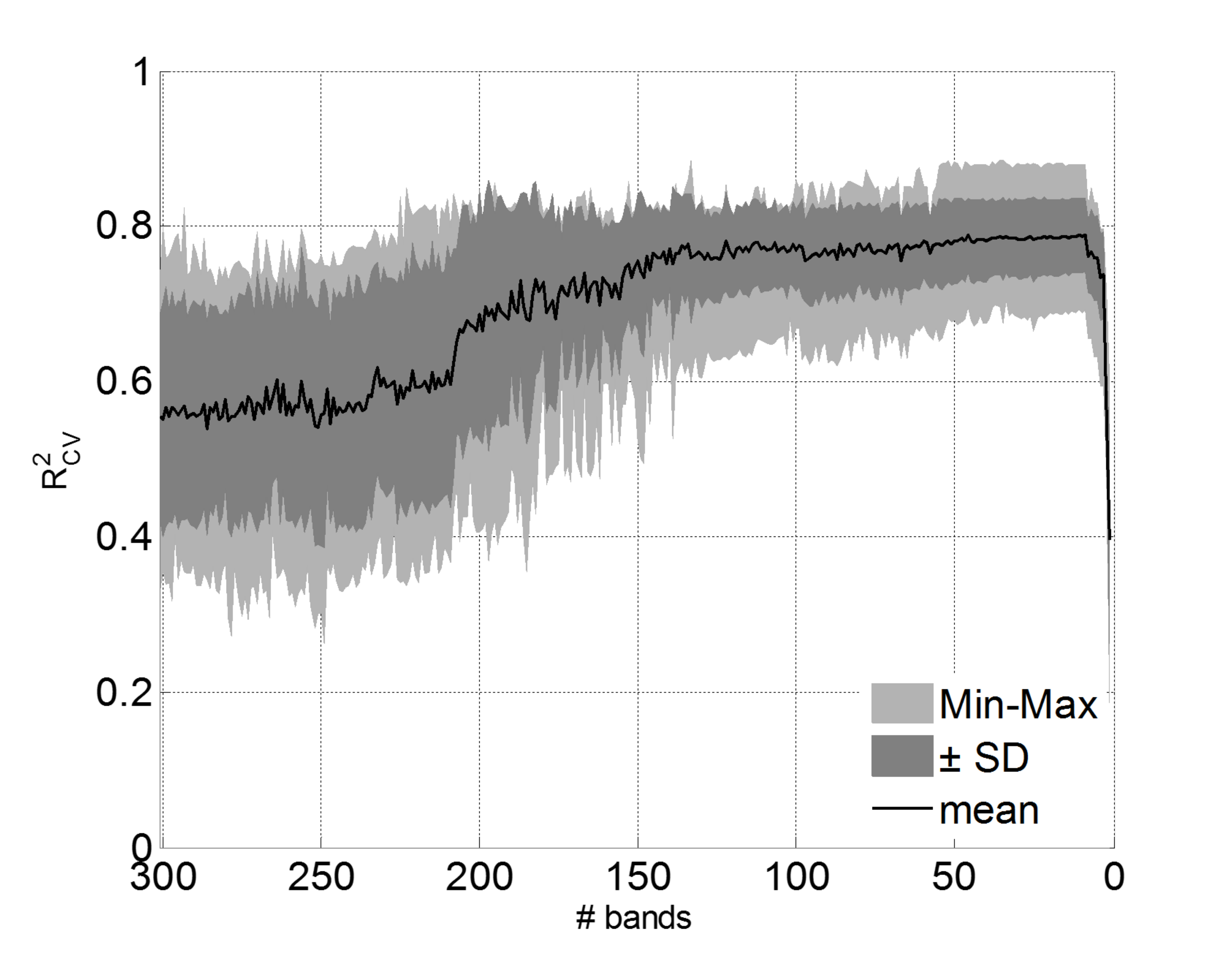} & 
\IG[width=6.0cm, trim={1cm 1cm 4cm 1cm}, clip]{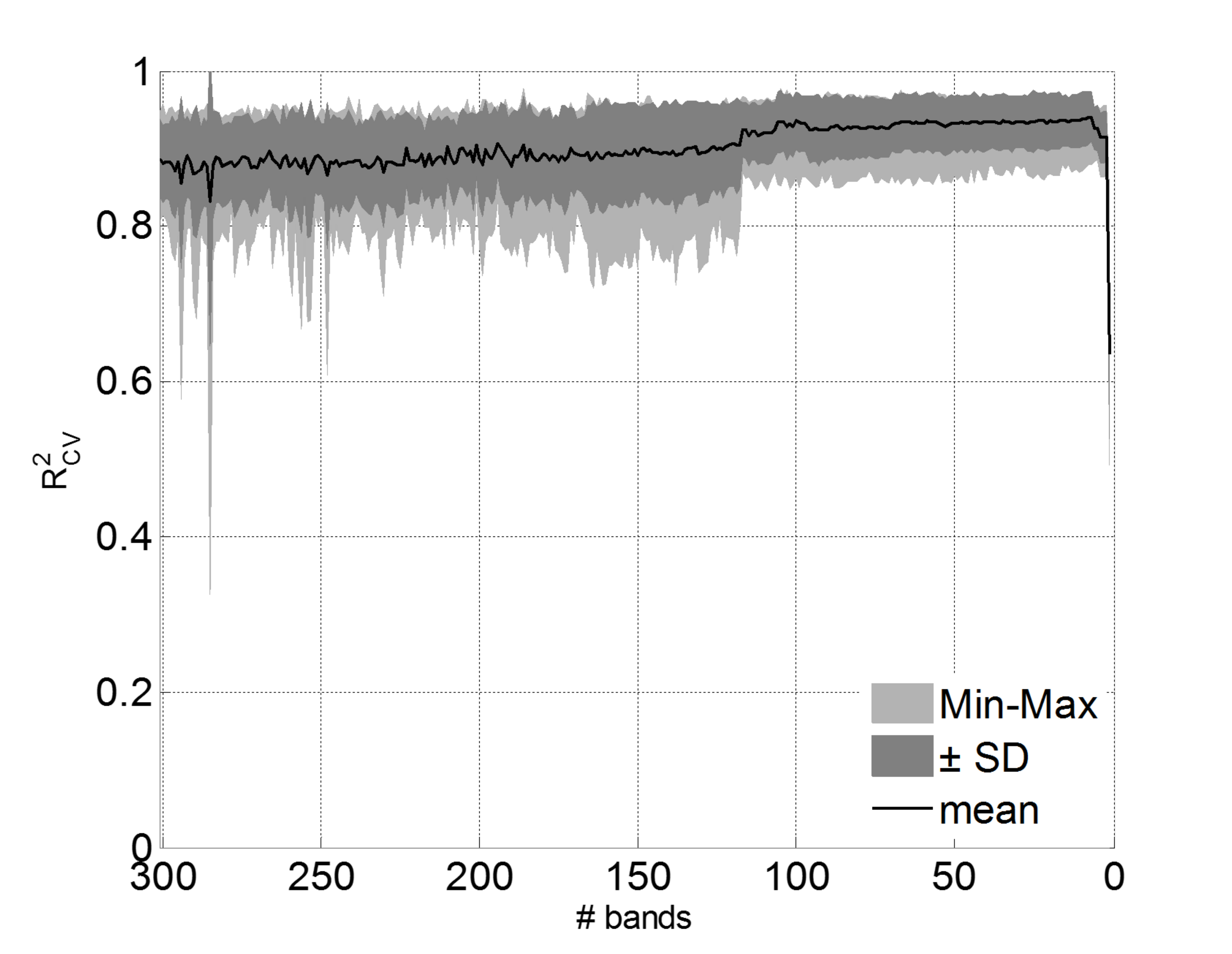}  \\
\IG[width=6.0cm, trim={1cm 1cm 4cm 1cm}, clip]{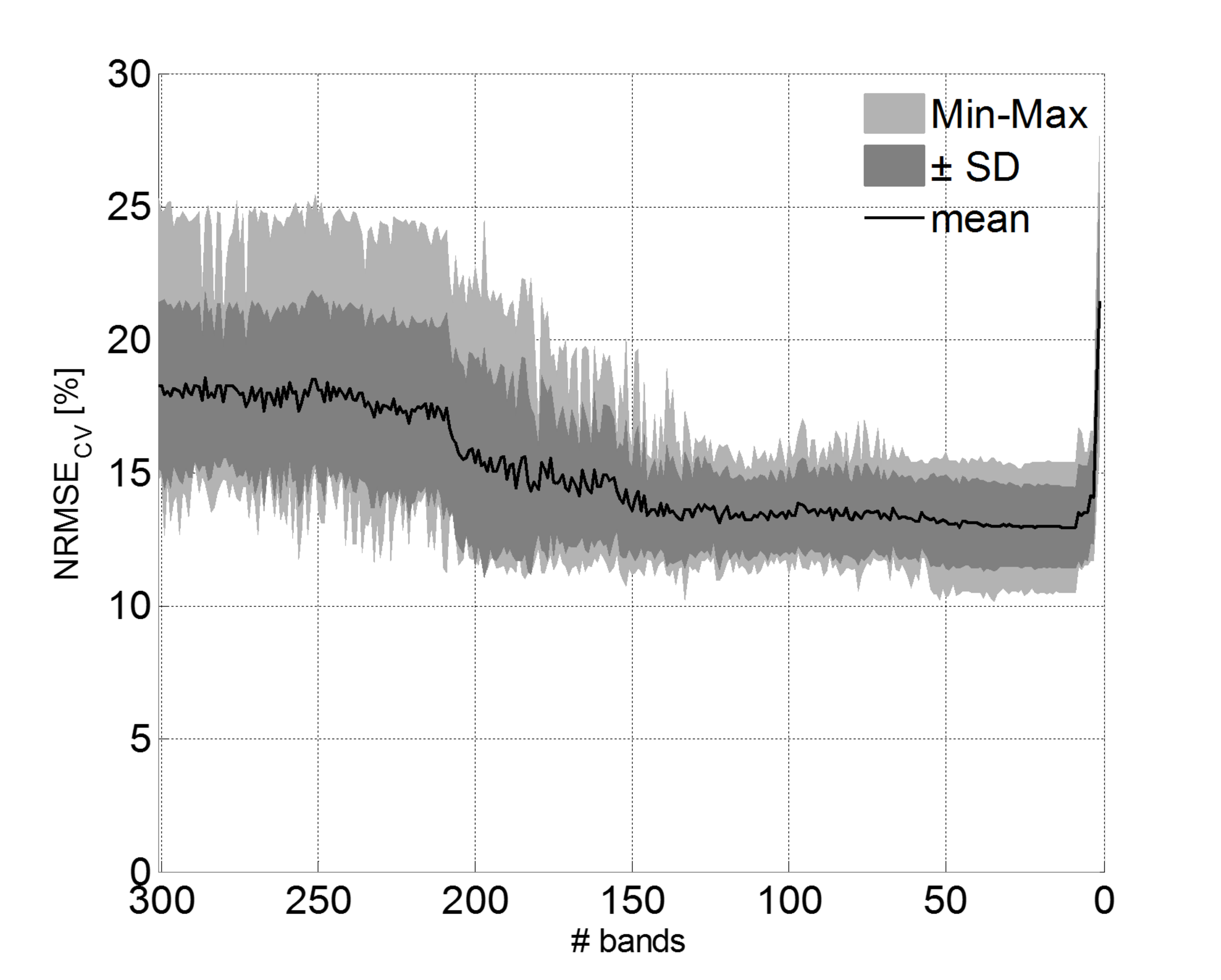} & 
\IG[width=6.0cm, trim={1cm 1cm 4cm 1cm}, clip]{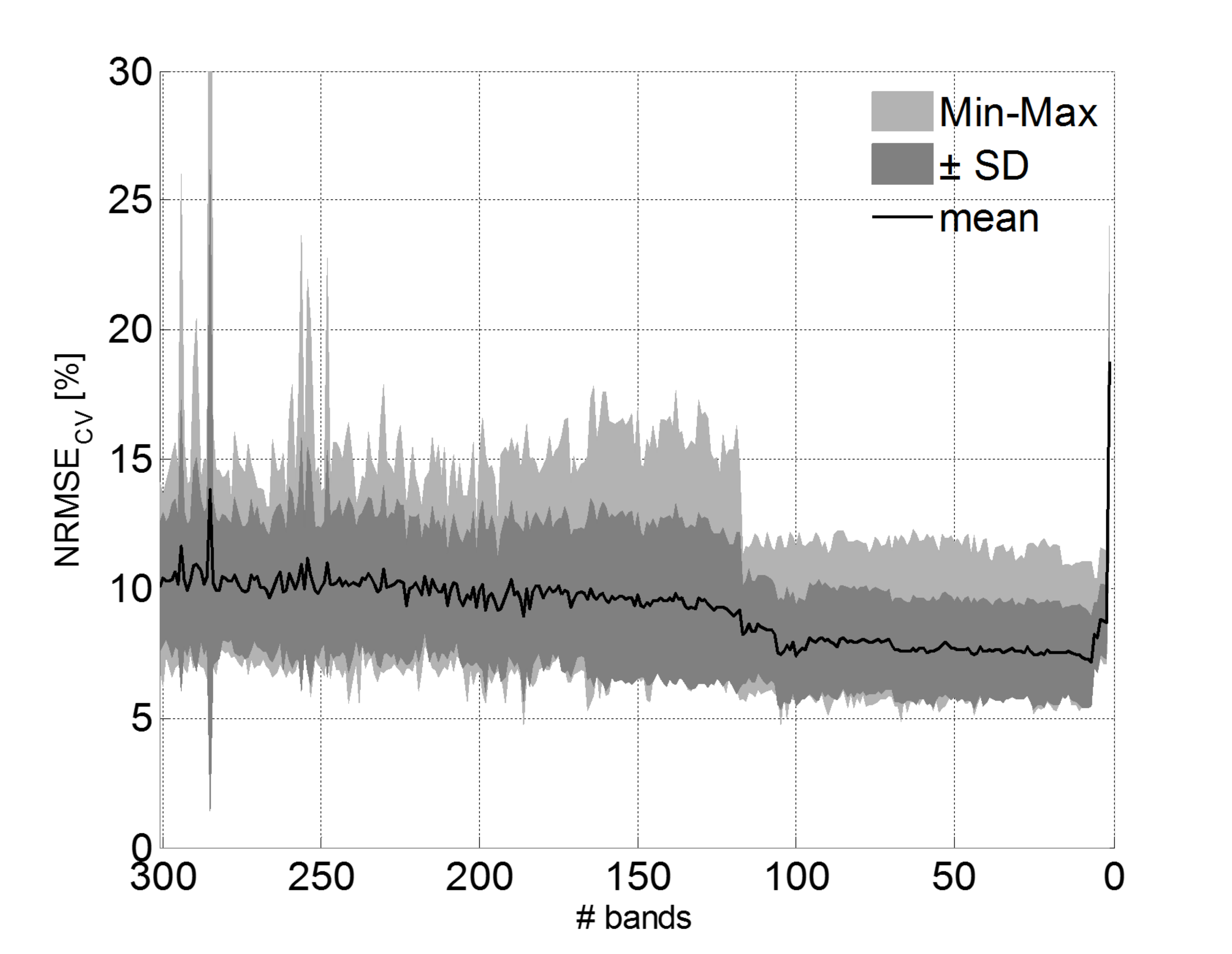}  \\
\end{tabular}
\end{center}
\caption{Cross-validation $R^2_{CV}$ (top) and NRMSE$_{CV}$ [\%] (bottom) statistics (mean, standard deviation, and min-max ranges) for LCC (left) and gLAI (right) plotted over sequentially removing the least contributing band.}
\label{original_LCC_gLAI_R2}
\end{figure}

The 10-fold CV SBBR analysis was initialized from the full-spectrum dataset until only one band was left. 
Averaged validation $R^2_{CV}$ results along with SD and min-max extremes for LCC and gLAI are provided in Figure \ref{original_LCC_gLAI_R2}. The $R^2_{CV}$ results of the final top 10 band combinations together with model processing time and the corresponding remaining wavelengths are listed in Table \ref{USB2000_GPR_wvl}. 

\begin{table}[t!] 
\begin{center}
\caption{Cross-validation $R^2_{CV}$ and NRMSE$_{CV}$ statistics (mean, standard deviation), processing time for a single model and associated wavelengths for LCC (top) and gLAI (bottom) according to iterative least contributing band removal for remaining top 10 bands. Additionally, results and processing time when using all bands is provided. \red{The best performing configuration is boldfaced.}} 
\vspace{-0.2cm}
\resizebox{1.0\textwidth}{!}{ 
\begin{tabular}{lllll}
\hline
\textbf{\# Bands} &  \textbf{$R^2_{CV}$ (SD)} & \textbf{NRMSE$_{CV}$ (SD)} &  \textbf{time (s)} & \textbf{Wavelengths (nm)}\\
\hline
\textbf{LCC} &&&&\\
\textit{301} & \textit{0.55 (0.13)} & \textit{ 18.25 (3.13)} & \textit{18.97} & \textit{all bands}\\
\vdots & \vdots & \vdots & \vdots & \vdots \\
10 & 0.79  (0.05) &  12.92 (1.52) &  1.05 & 482, 500, 564, 566, 710, 712, 714, 878, 966, 980 \\
\textbf{9} & \textbf{0.79 (0.05) }&  \textbf{12.90 (1.53)} &  \textbf{1.01} & \textbf{482, 500, 564, 710, 712, 714, 878, 966, 980} \\ 
 8 & 0.76  (0.05) &  13.51 (1.82) &  0.85 & 482, 500, 564, 710, 712, 714, 878, 966\\
 7 & 0.77  (0.06) &  13.35 (1.92) &  0.87 & 482, 500, 564, 710, 714, 878, 966 \\
 6 & 0.76  (0.05) &  13.51 (1.75) &  0.83 & 482, 500, 710, 714, 878, 966\\
 5 & 0.76  (0.06) &  13.48 (1.78) &  0.77 & 500, 710, 714, 878, 966\\ 
 4 & 0.73  (0.06) &  14.10 (1.74) &  0.69 & 500, 710, 714, 878\\ 
 3 & 0.74  (0.06) &  14.04 (1.71) &  0.64 & 500, 710, 878\\ 
 2 & 0.56  (0.06) &  18.24 (2.62) &  0.59 & 500, 710\\ 
 1 & 0.40  (0.13) &  21.42 (4.04) &  0.50 & 710\\ 
\\
\textbf{gLAI} &&&&\\
\textit{301} & \textit{0.88 (0.12)} & \textit{10.07 (2.48)}  & \textit{18.90}  & \textit{all bands}\\
\vdots & \vdots & \vdots& \vdots & \vdots \\
10 & 0.94  (0.04) & 7.34 (1.93) & 1.09 & 406, 746, 770, 790, 792, 794, 798, 808, 858, 878\\
 9 & 0.94  (0.04) & 7.30 (1.91) & 0.99 & 406, 746, 790, 792, 794, 798, 808, 858, 878 \\
 8 & 0.94  (0.03) & 7.27 (1.86) & 0.92 & 406, 746, 790, 792, 794, 798, 858, 878 \\ 
 \textbf{7} & \textbf{0.94 (0.03)} & \textbf{7.19 (1.76)} & \textbf{0.86} & \textbf{406, 746, 792, 794, 798, 858, 878} \\  
 6 & 0.93  (0.03) & 8.19 (1.26) & 0.82 & 746, 792, 794, 798, 858, 878 \\
 5 & 0.93  (0.03) & 8.12 (1.35) & 0.77 & 746, 792, 794, 798, 878\\ 
 4 & 0.91  (0.03) & 8.81 (1.37) & 0.70 & 746, 792, 794, 798\\ 
 3 & 0.91  (0.03) & 8.75 (1.43) & 0.64 & 746, 792, 794 \\  
 2 & 0.92  (0.03) & 8.72 (1.42) & 0.57 & 746, 792 \\  
 1 & 0.64  (0.11) & 18.77 (3.49) & 0.51 & 792\\ 
\hline
\hline
\label{USB2000_GPR_wvl}
\end{tabular}}
\end{center}
\end{table}

These results lead to the following main findings:
\begin{itemize}
	\item Applying all bands into the GPR model did not lead to the best performance. Results were more unstable (see large SD, min-max) and poorer than in case of e.g. using only 50 to 3 best bands and processing time was significantly larger, i.e. 19 s for a model using hyperspectral data as opposed to less than a second in case of using less than 10 remaining bands. Especially, LCC significantly gained from band removal, e.g. performances systematically improved when having about 150 bands removed. Also, when using less bands, the likelihood of developing extremely poor models (see the minima in light gray) decreased. 
	\item When restricting to about the remaining 50 bands, validation results kept stable until relatively few bands remained. For LCC stable results maintained until 9 bands are left while for gLAI even maintained the same high accuracies until only 7 bands are left. Thereby, accuracies kept high until three bands (LCC) and two bands (gLAI) remained. In turn, using only one band led to worst result - the only spectral configuration that performed significantly worse than when using all bands.
	\item Overall, LCC benefited most from bands around the \red{blue (482 nm)} green peak (500, 564 nm) and in the red edge (710, 714 nm), and subsequently from bands in the NIR region (878-980 nm) when adding more bands. gLAI benefited most from a band in the red edge (746 nm) and bands in the NIR (792-878 nm) and a band in the blue (406 nm). 
\end{itemize}

\begin{figure}[h!]
\begin{center}
\begin{tabular}{cc}
LCC & gLAI \\
\IG[width=6.0cm, trim={0.5cm 0.0cm 0.5cm 0.5cm}, clip]{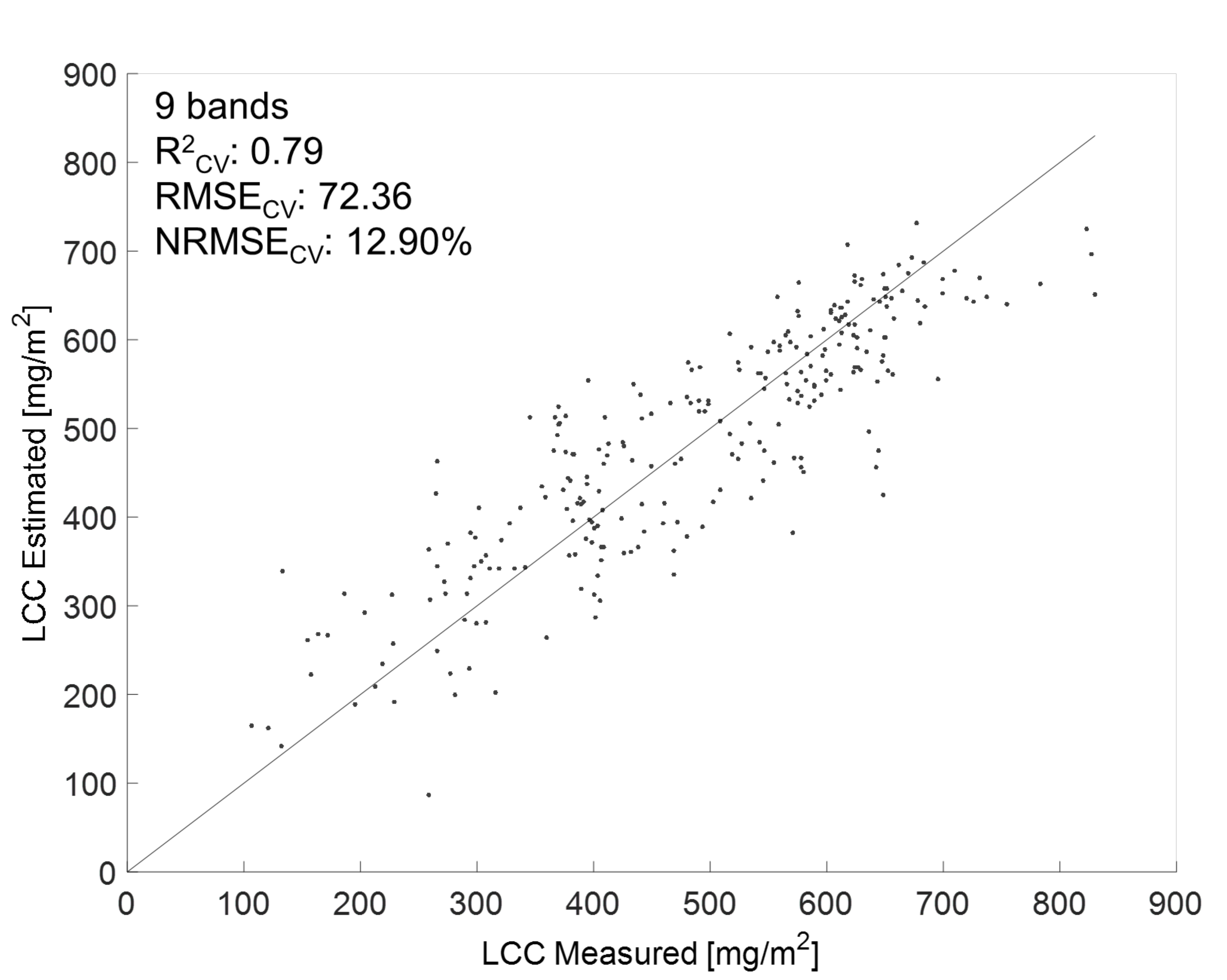} &
\IG[width=6.0cm, trim={0.5cm 0.0cm 0.0cm 0.5cm}, clip]{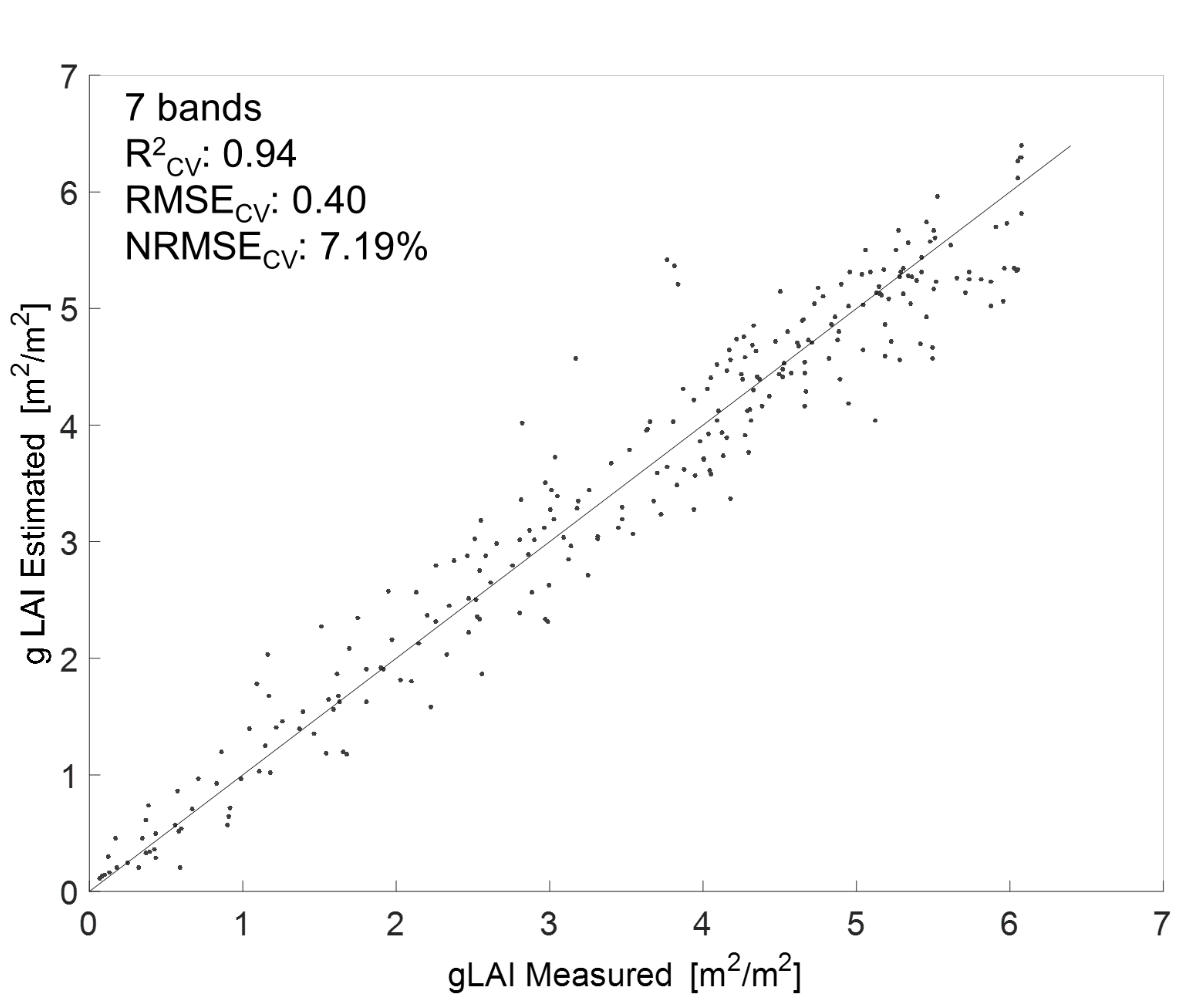} \\
\end{tabular}
\end{center}
\caption{Measured vs estimated LCC (left) and gLAI (right) values along the 1:1-line of the best performing GPR model (see Table \ref{USB2000_GPR_wvl}).}
\label{USB2000_meas_est}
\end{figure}

For the best performing models for LCC (9 bands) and gLAI (7 bands) a scatterplot of estimations against measured data is shown in Figure \ref{USB2000_meas_est}. It can be noted that particularly gLAI is well estimated, with the large majority of samples closely located around the 1:1-line. The NRMSE$_{CV}$ is well below the 10\%, which is commonly regarded as accuracy threshold by end users~\citep{Drusch2012}.

\subsection{Barrax SPARC dataset: LAI and CWC retrieval}

\begin{figure}[h!]
\begin{center}
\begin{tabular}{cc}
LAI & CWC \\
\IG[width=6.0cm, trim={1cm 1cm 4cm 1cm}, clip]{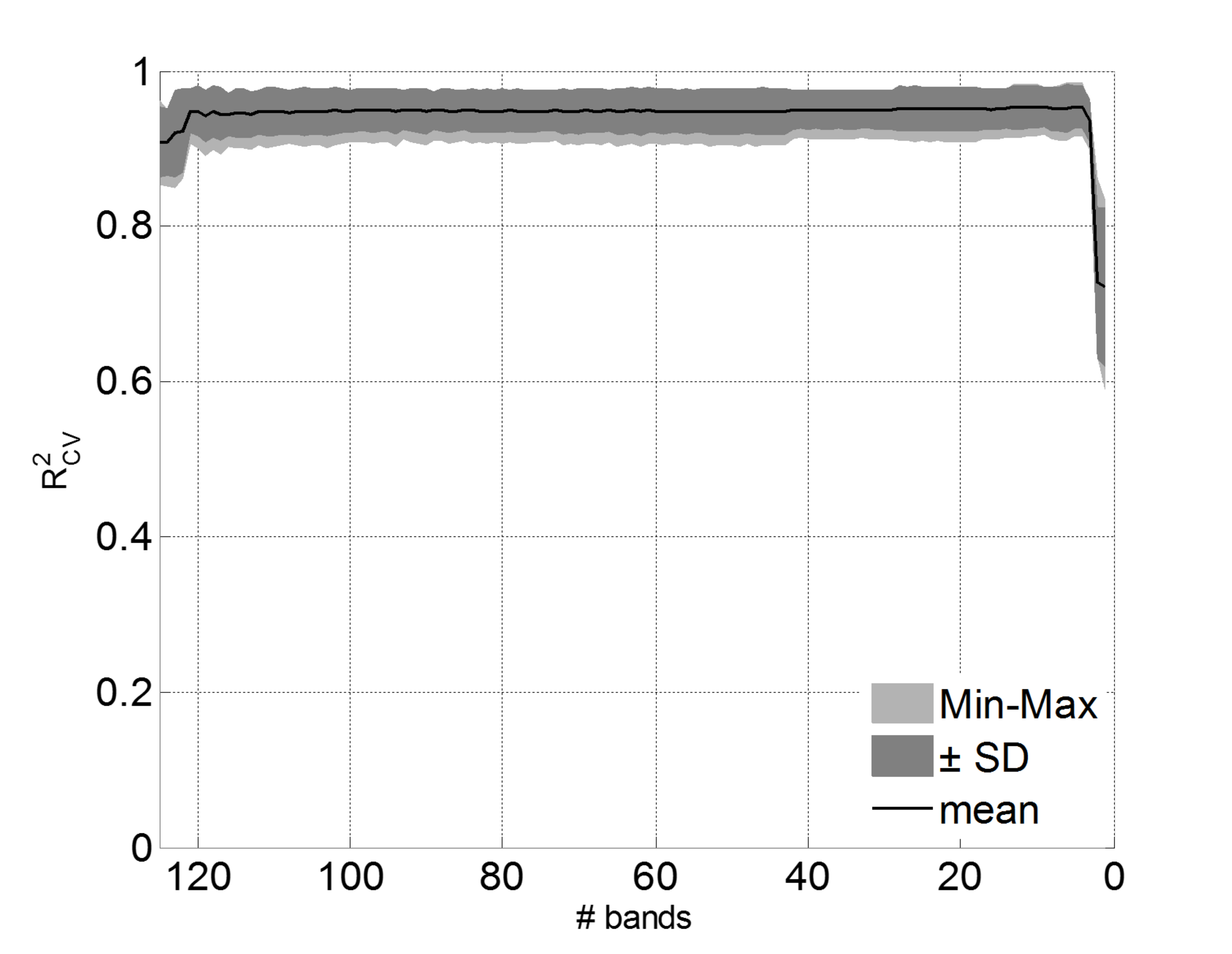} & 
\IG[width=6.0cm, trim={1cm 1cm 4cm 1cm}, clip]{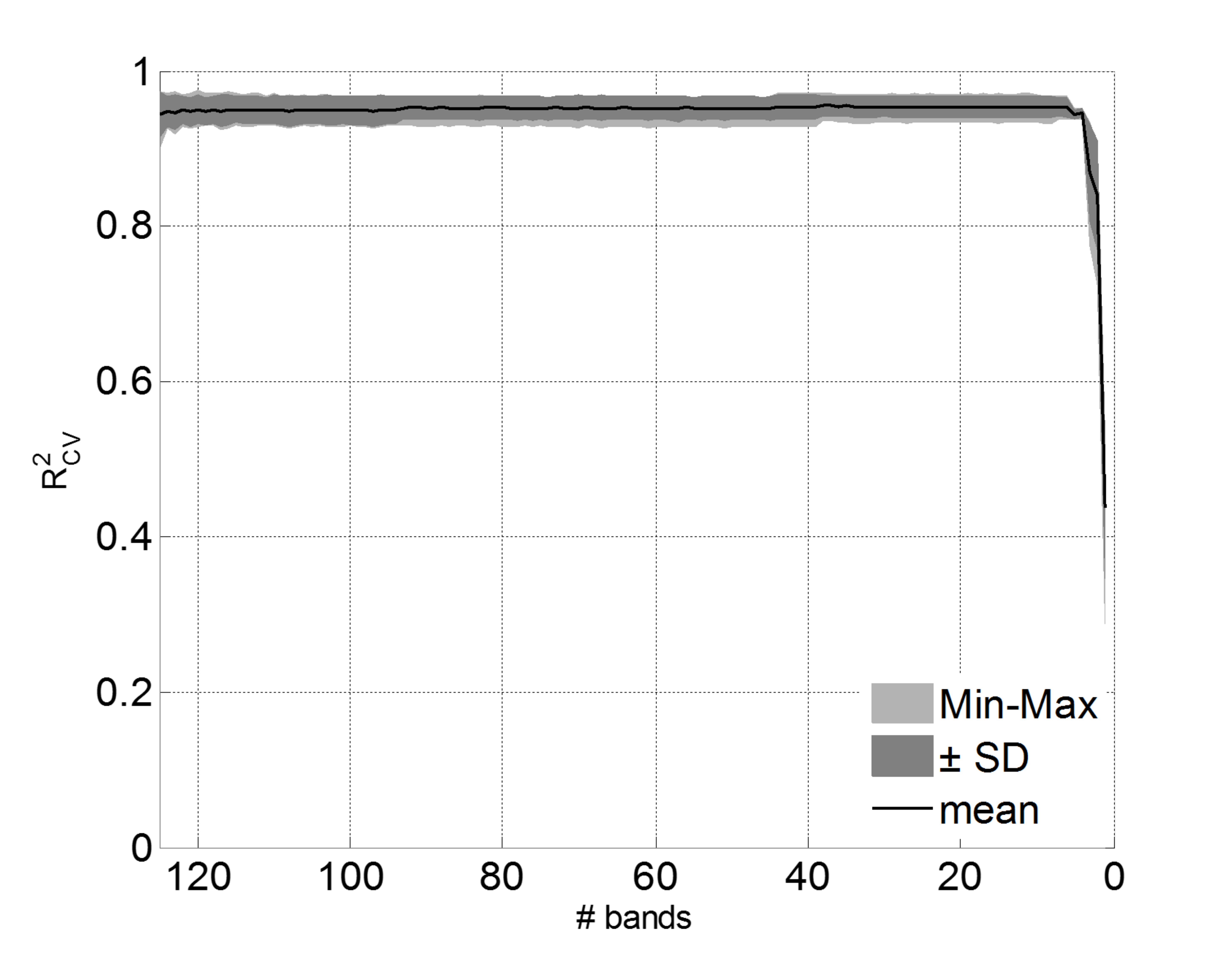}  \\
\IG[width=6.0cm, trim={1cm 1cm 4cm 1cm}, clip]{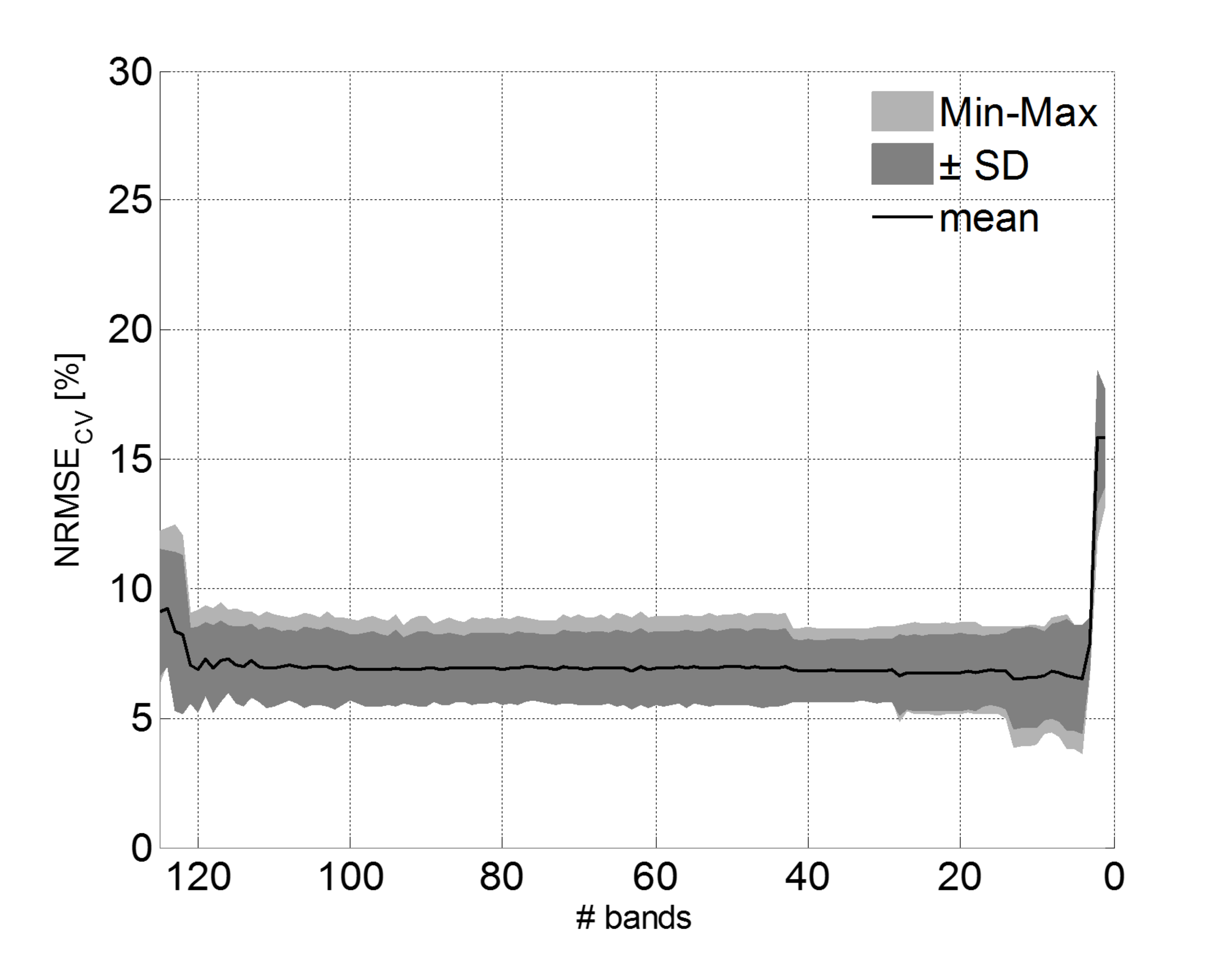} & 
\IG[width=6.0cm, trim={1cm 1cm 4cm 1cm}, clip]{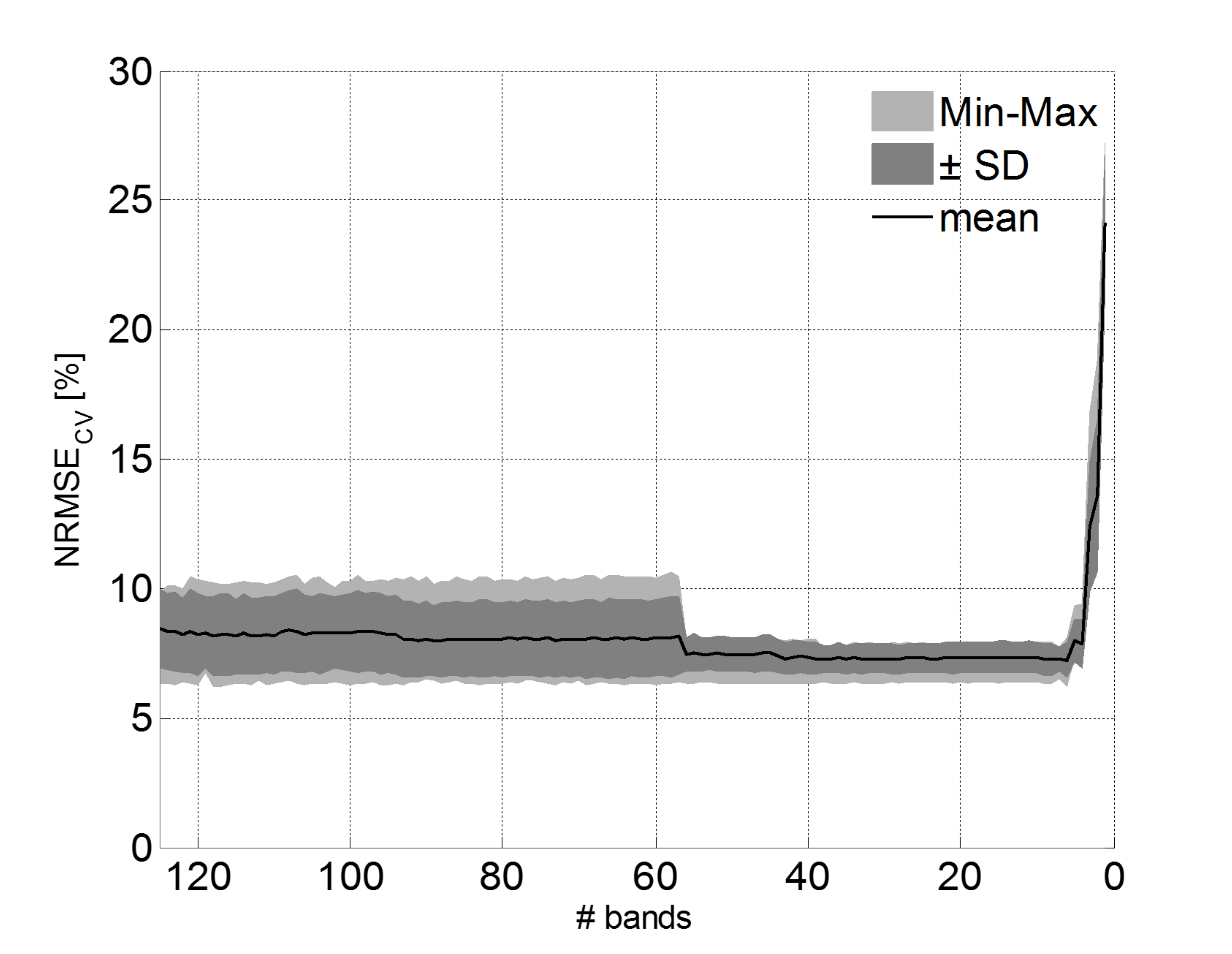}  \\
\end{tabular}
\end{center}
\caption{Cross-validation $R^2_{CV}$ (top) and NRMSE$_{CV}$ [\%] (bottom) statistics (mean, standard deviation, and min-max ranges) for LAI (left) and CWC (right) plotted over sequentially removing the least contributing band.}
\label{SPARC_LAI_CWC_R2}
\end{figure}

Next, a 4-fold CV SBBR analysis was applied to the SPARC dataset, acquired at Barrax, Spain. The spectral data comes from the airborne HyMap image, which was configured with 125 bands. This is considerably less than the field hyperspectral dataset and causes that spectral redundancy is less an issue. That is also observable in the plotted averaged validation $R^2_{CV}$ and NRMSE$_{CV}$ results along with SD and min-max extremes (Figure \ref{SPARC_LAI_CWC_R2}). Also less samples are included (\#100), which implies faster training and validation of the GPR models. The performances of the last 10 bands and processing time are shown in Table \ref{HyMap_GPR_wvl}.

These results lead to the following main findings:
\begin{itemize}
	\item Contrary to the earlier field hyperspectral dataset, accuaries kept stable from the initial 125 bands until only a few bands are left. Only, regarding the LAI dataset, it can be noted that results are suboptimal when using all bands; by reducing to 120 bands performances stabilize. While the mean $R^2_{CV}$ results of CWC kept stable until a few bands are left, when inspecting the mean NRMSE$_{CV}$, results significantly improved and kept more stable when reducing to 60 bands until a few bands are left. Also processing time for generating a GPR model improved from about 4 seconds (all bands) until less than half a second when restricting to less than 10 bands. 
	\item	For LAI stable results maintained until 4bands are left while for CWC  the same high accuracies maintained until  6 bands are left. Thereby, accuracies kept high until three bands (LAI) and four bands (CWC) remained. In turn, using only one band led to worst result.
	\item Overall, LAI benefited most from a band in the blue (462 nm) bands in the red edge (708, 723 nm) and a band in the NIR (1327 nm). CWC benefited most from a band in the red edge (723 nm) and bands in the NIR (1157-1419 nm).
\end{itemize}

\begin{table}[t!] 
\begin{center}
\caption{Cross-validation $R^2_{CV}$ and NRMSE$_{CV}$ statistics (mean, standard deviation), processing time for a single model and associated wavelengths for LAI (top) and CWC (bottom) according to iterative least contributing band removal for remaining top 10 bands. Additionally, results and processing time when using all bands is provided. \red{The best performing configuration is boldfaced.}} 
\vspace{-0.2cm}
\resizebox{1.0\textwidth}{!}{ 
\begin{tabular}{lllll}
\hline
\textbf{\# Bands} &  \textbf{$R^2_{CV}$ (SD)} & \textbf{NRMSE$_{CV}$ (SD)} &  \textbf{time (s)} & \textbf{Wavelengths (nm)}\\
\hline
\textbf{LAI} &&&&\\
\textit{125} & \textit{0.91 (0.05)} & \textit{9.10 (2.44)} & \textit{3.94} & \textit{all bands}\\
\vdots & \vdots & \vdots & \vdots & \vdots \\
10 & 0.95  (0.03)  &  6.55 (1.92)   & 0.48 & 462, 478, 708, 723, 1215, 1243, 1272, 1327, 1635, 2483 \\
 9 & 0.95  (0.03)  &  6.65 (1.72)  & 0.44 & 462, 478, 708, 723, 1215, 1243, 1272, 1327, 2483\\
 8 & 0.95  (0.03)  &  6.79 (1.82)   & 0.45 & 462, 478, 708, 723, 1215, 1243, 1272, 1327\\
 7 & 0.95  (0.03)  &  6.76 (2.14)   & 0.41 & 462, 478, 708, 723, 1215, 1272, 1327\\
 6 & 0.95  (0.03)  &  6.65 (2.14)   & 0.35 & 462, 478, 708, 723, 1215, 1327\\
 5 & 0.95  (0.03)  &  6.55 (2.01)   & 0.31 & 462, 478, 708, 723, 1327\\
 \textbf{4} & \textbf{0.95 (0.03)}  & \textbf{6.50 (2.09)} & \textbf{0.28} & \textbf{462, 708, 723, 1327}\\ 
 3 & 0.94  (0.03)  &  7.88 (1.02)  & 0.26 & 462, 708, 1327\\
 2 & 0.73  (0.10)  &  15.81 (2.60) & 0.22 & 462, 1327\\
 1 & 0.72  (0.10)  &  15.85 (1.89)  & 0.19 & 462\\ 
\\
\textbf{CWC} &&&&\\
\textit{125} & \textit{0.94 (0.03)} & \textit{8.46 (1.54} & \textit{3.94} & \textit{all bands}\\
\vdots & \vdots & \vdots & \vdots & \vdots \\
10 & 0.95  (0.01)  &  7.33 (0.59)  & 0.45 & 462, 723, 1128, 1157, 1272, 1286, 1299, 1327, 1419, 2483 \\
 9 & 0.95  (0.01)  &  7.27 (0.61)  & 0.45 & 723, 1128, 1157, 1272, 1286, 1299, 1327, 1419, 2483\\
 8 & 0.95  (0.01)  &  7.27 (0.61)  & 0.40 & 723, 1128, 1157, 1272, 1286, 1327, 1419, 2483\\
 7 & 0.95  (0.01)  &  7.28 (0.47)  & 0.40 & 723, 1128, 1157, 1272, 1286, 1327, 1419\\
 \textbf{6} & \textbf{0.95 (0.01)}  & \textbf{7.24 (0.67)}  & \textbf{0.34} & \textbf{723, 1157, 1272, 1286, 1327, 1419}\\ 
 5 & 0.94  (0.01)  &  7.98 (0.80)  & 0.31 & 723, 1157, 1272, 1286, 1327\\
 4 & 0.95  (0.01)  &  7.88 (0.94)   & 0.28 & 723, 1157, 1272, 1286\\
 3 & 0.87  (0.06)  &  12.39 (2.50)   & 0.25 & 1157, 1272, 1286\\
 2 & 0.84  (0.07)  &  13.56 (2.94)   & 0.21 & 1157, 1286\\
 1 & 0.44  (0.09)  &  24.15 (2.65)  & 0.19 & 1286\\ 
\hline
\label{HyMap_GPR_wvl}
\end{tabular}}
\end{center}
\end{table}

\begin{figure}[h!]
\begin{center}
\begin{tabular}{cc}
LAI & CWC \\
\IG[width=6.0cm, trim={0.5cm 0.0cm 0.5cm 0.5cm}, clip]{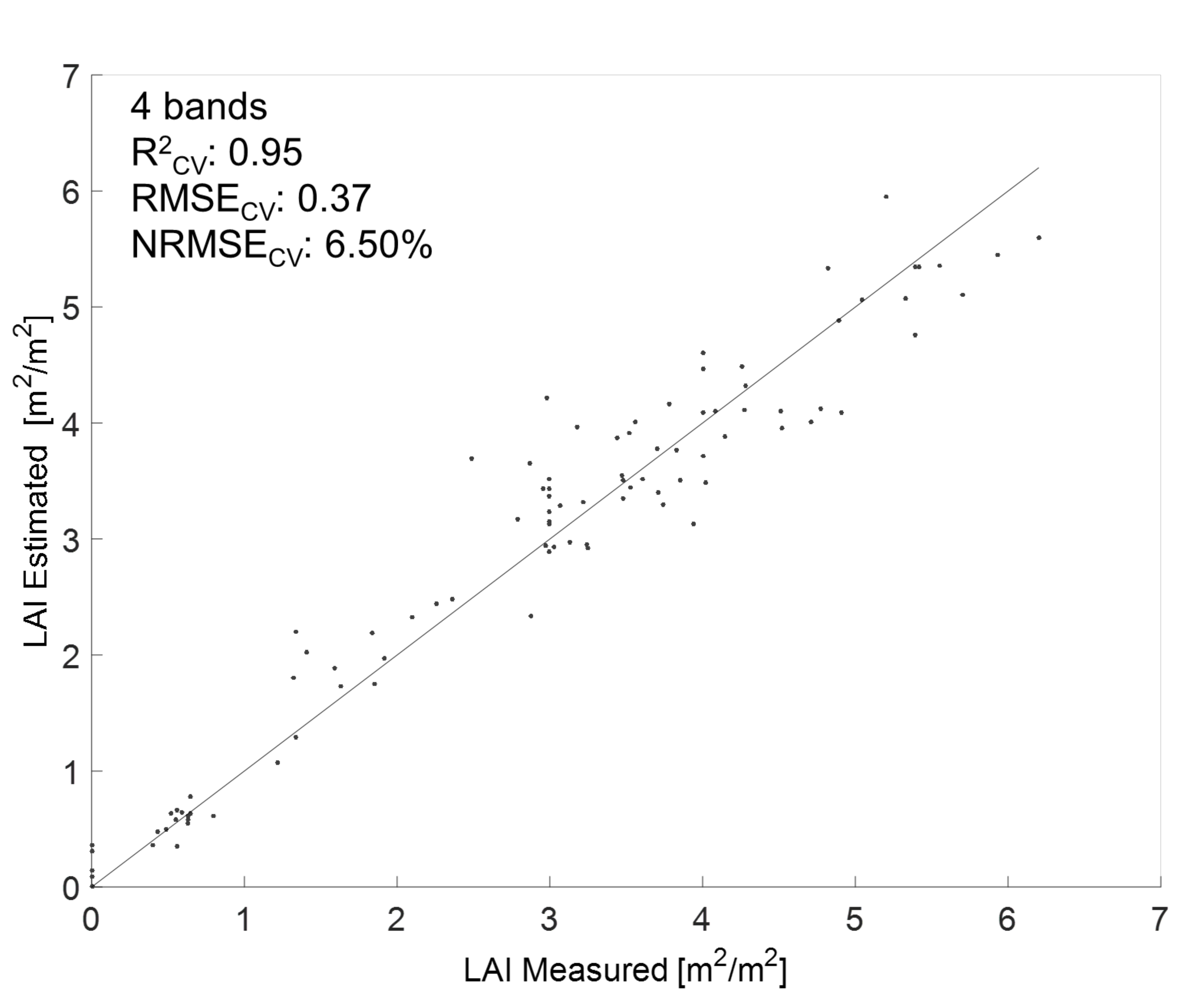} & %
\IG[width=6.0cm, trim={0.5cm 0.0cm 0.0cm 0.5cm}, clip]{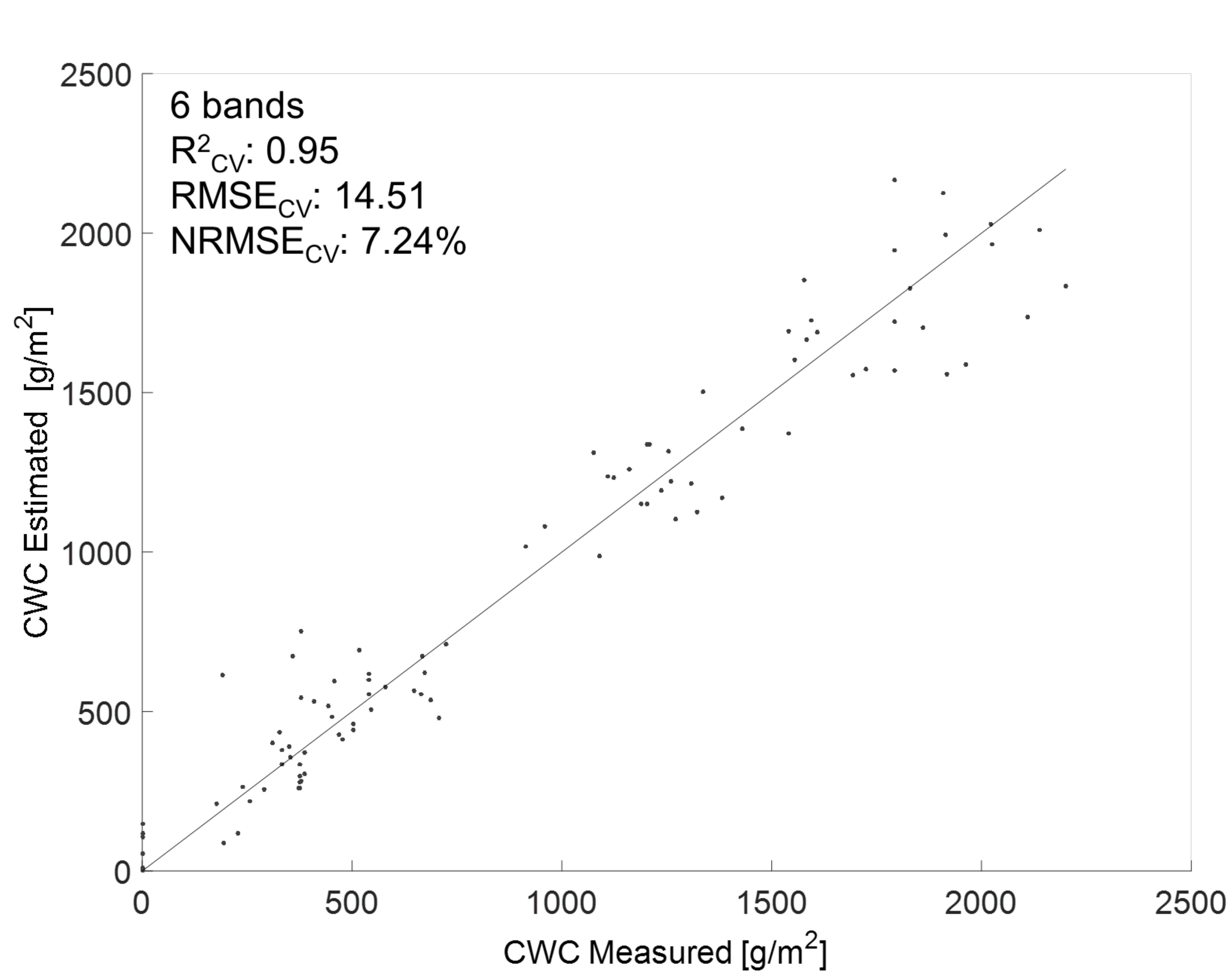} \\ 
\end{tabular}
\end{center}
\caption{Measured vs estimated LAI (left) and CWC (right) values along the 1:1-line of the best performing GPR model (see Table \ref{HyMap_GPR_wvl}).}
\label{HyMap_meas_est}
\end{figure}

For the best performing models for LAI (4 bands) and CWC (6 bands) a scatterplot of estimations against measured data is shown in Figure \ref{HyMap_meas_est}. It can be noted that both biophysical variables are well estimated, with the large majority of samples closely located around the 1:1-line. The NRMSE$_{CV}$ is well below the 10\%, which is commonly regarded as accuracy threshold by end users~\citep{Drusch2012}. These are thus both valid retrieval models, with the limitation that the SPARC dataset is considerably smaller than the UNL field hyperspectral dataset.

\begin{figure}[t!]
\begin{center}
\begin{tabular}{ccc}
LAI $\mu$ [m$^2$/m$^2$] & LAI $\sigma$ [m$^2$/m$^2$]& LAI CV [\%]\\
\IG[width=2cm, height=7cm, trim={0.cm 0.0cm 0.cm 0.5cm}, clip]{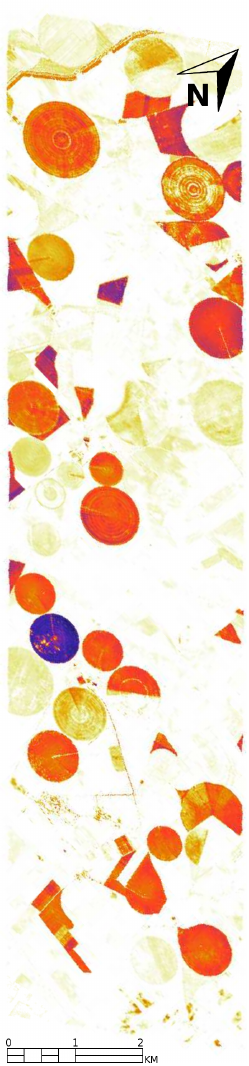} & %
\IG[width=2cm, height=7cm, trim={0cm 0.0cm 0cm 0.5cm}, clip]{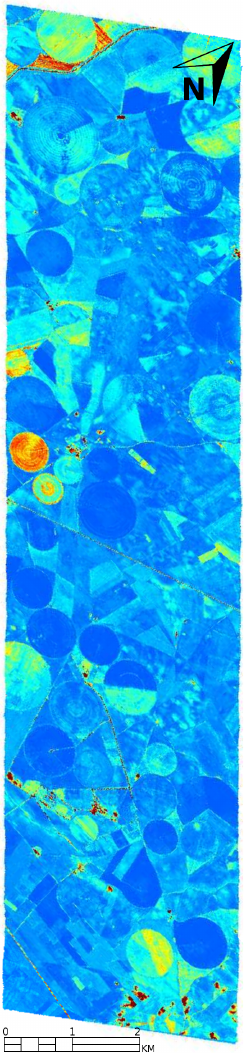} &
\IG[width=2cm, height=7cm, trim={0cm 0.0cm 0cm 0.5cm}, clip]{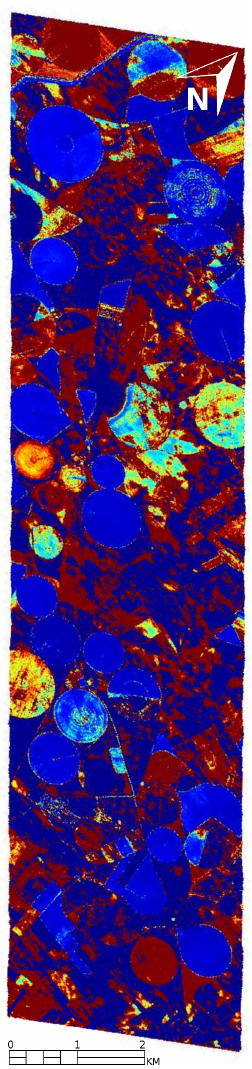}\\ 
\IG[width=2.2cm, height=0.5cm]{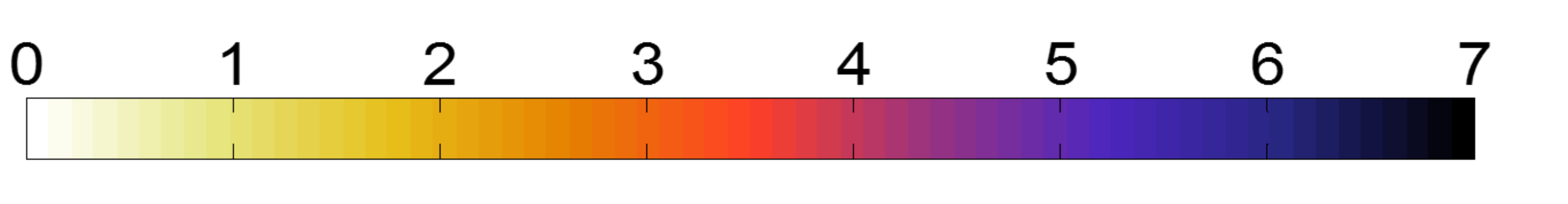} & %
\IG[width=2.2cm, height=0.5cm]{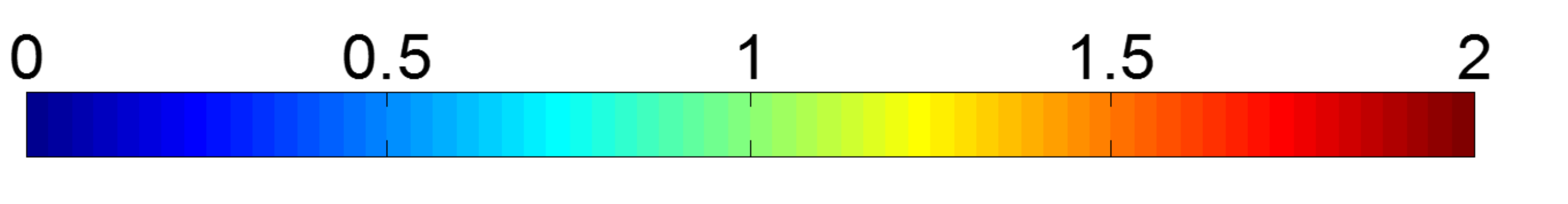} &
\IG[width=2.2cm, height=0.5cm]{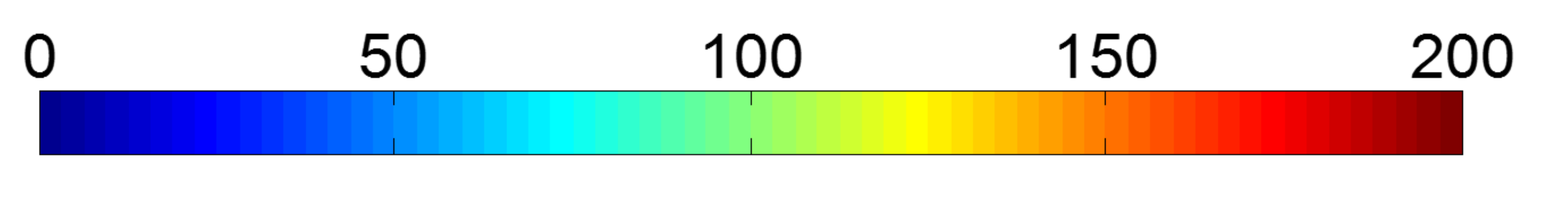}\\ 
CWC $\mu$ [g/m$^2$] & CWC $\sigma$ [g/m$^2$] & CWC CV [\%]\\
\IG[width=2cm, height=7cm, trim={0cm 0.0cm 0cm 0.5cm}, clip]{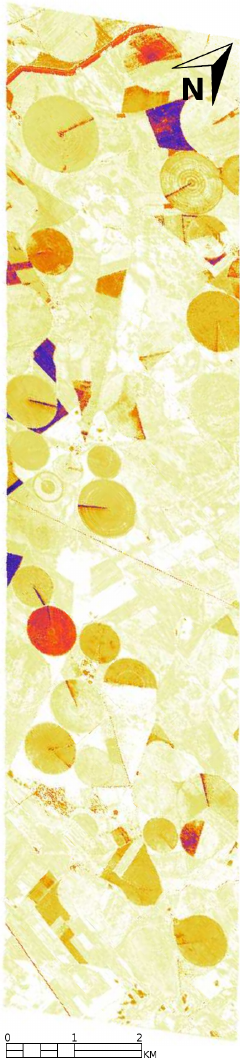} & %
\IG[width=2cm, height=7cm, trim={0cm 0.0cm 0cm 0.5cm}, clip]{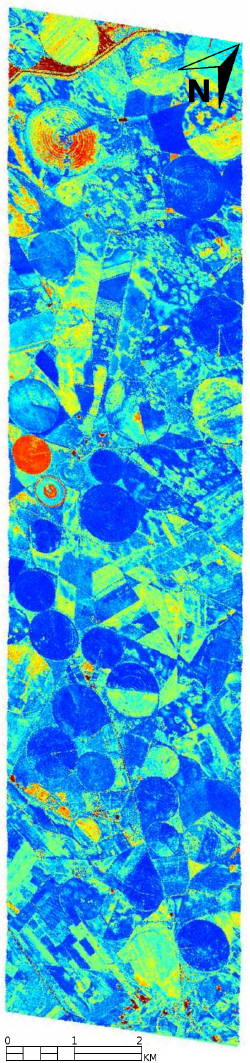} &
\IG[width=2cm, height=7cm, trim={0cm 0.0cm 0cm 0.5cm}, clip]{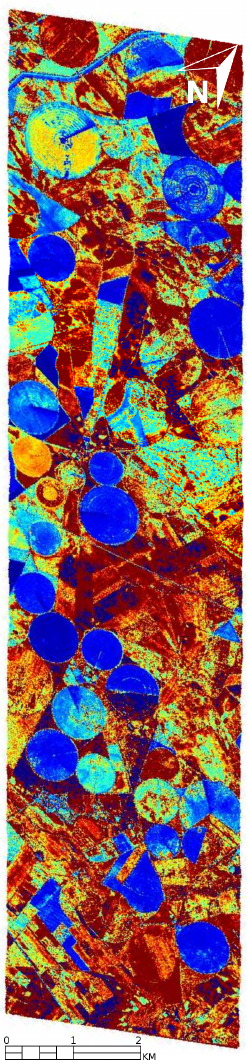}\\ 
\IG[width=2.2cm, height=0.5cm]{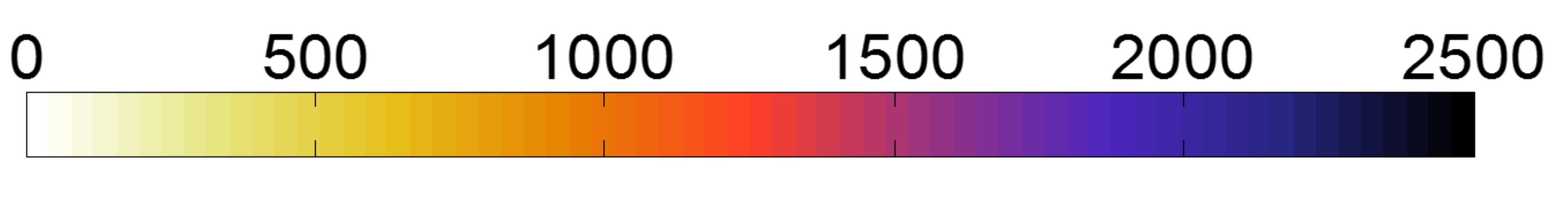} & %
\IG[width=2.2cm, height=0.5cm]{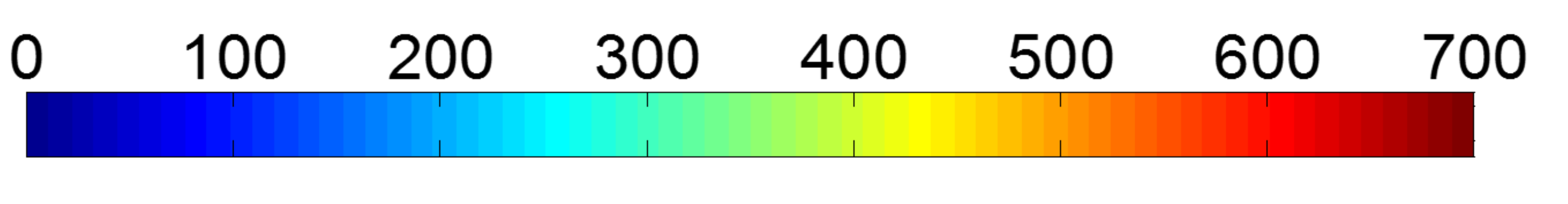} &
\IG[width=2.2cm, height=0.5cm]{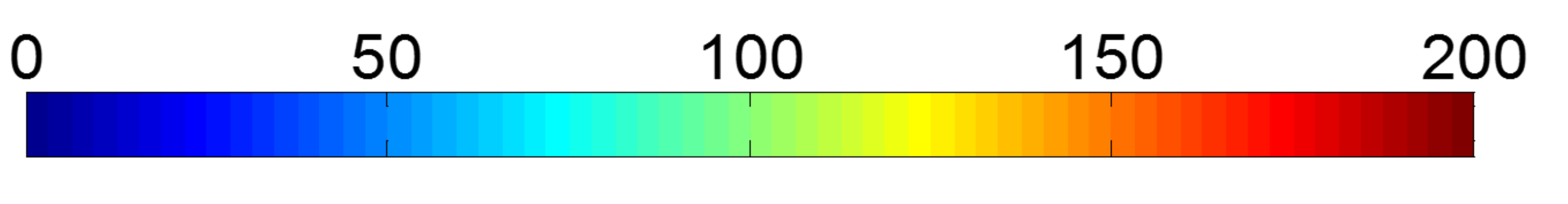}\\ 

\end{tabular}
\end{center}
\caption{HyMap LAI (top) and CWC (bottom) maps: mean estimates; $\mu$) (left), associated uncertainties $\sigma$ (expressed as standard deviation around the $\mu$) (centre), and relative uncertainties (expressed as coefficient of variation: CV= $\sigma$/$\mu$ *100 [\%]) (right) as generated by the best performing GPR model (see Table \ref{HyMap_GPR_wvl}).}
\label{HyMap_maps}
\end{figure}

Given these optimized regression models, they were subsequently applied to an HyMap flight line. The toolbox directly selects the right bands within the image and processes it into the targeted vegetation property. The mean prediction map of LAI and CWC is provided Figure \ref{HyMap_maps} (left). 
Moreover, GPR-BAT can be used in combination with other advantages of GPR for mapping vegetation products. Since GPR models yield a full posterior predictive distribution, it is possible to map not only mean predictions, but also the so-called ``error-bars'' ($\sigma$), i.e. the uncertainty of the mean prediction~\citep{Verrelst2012,Verrelst12a,Verrelst13a,Verrelst2013b}. These uncertainty maps are provide in Figure \ref{HyMap_maps} (middle). It should be kept in mind that $\sigma$ is also related to the magnitude of the mean estimates ($\mu$). For this reason relative uncertainties ($\sigma$/$\mu$) may provide a more meaningful interpretation. Those maps are also shown in Figure \ref{HyMap_maps} (right). It can be observed that LAI and CWC are retrieved with a high relative certainty over the \red{center-pivot irrigation crop circles}. Typically, relative uncertainties below 20\% are achieved for several of those areas, which falls within the accuracy threshold as proposed by Global Climate Observing System (GCOS) ~\citep{GCOS11}. Note that on the fallow areas or bare soils, retrievals have a rather high relative uncertainty. The high uncertainties is due to that the training data is almost exclusively coming from samples collected within the \red{center-pivot irrigation circles}, which consists of vegetative crops. By applying a threshold those more uncertain retrievals can be masked out. Hence, uncertainty maps can function as a spatial mask that enables displaying only pixels with a high certainty. All in all, with GPR-BAT best performing bands can be identified based on training data, while with the associated uncertainty estimates the GPR retrieval performance over an entire image can be assessed.

\subsection{Interpretation of sensitive bands}

\begin{figure}[t!]
\begin{center}
\IG[width=13.0cm, trim={1cm 0.5cm 0.0cm 0.5cm}, clip]{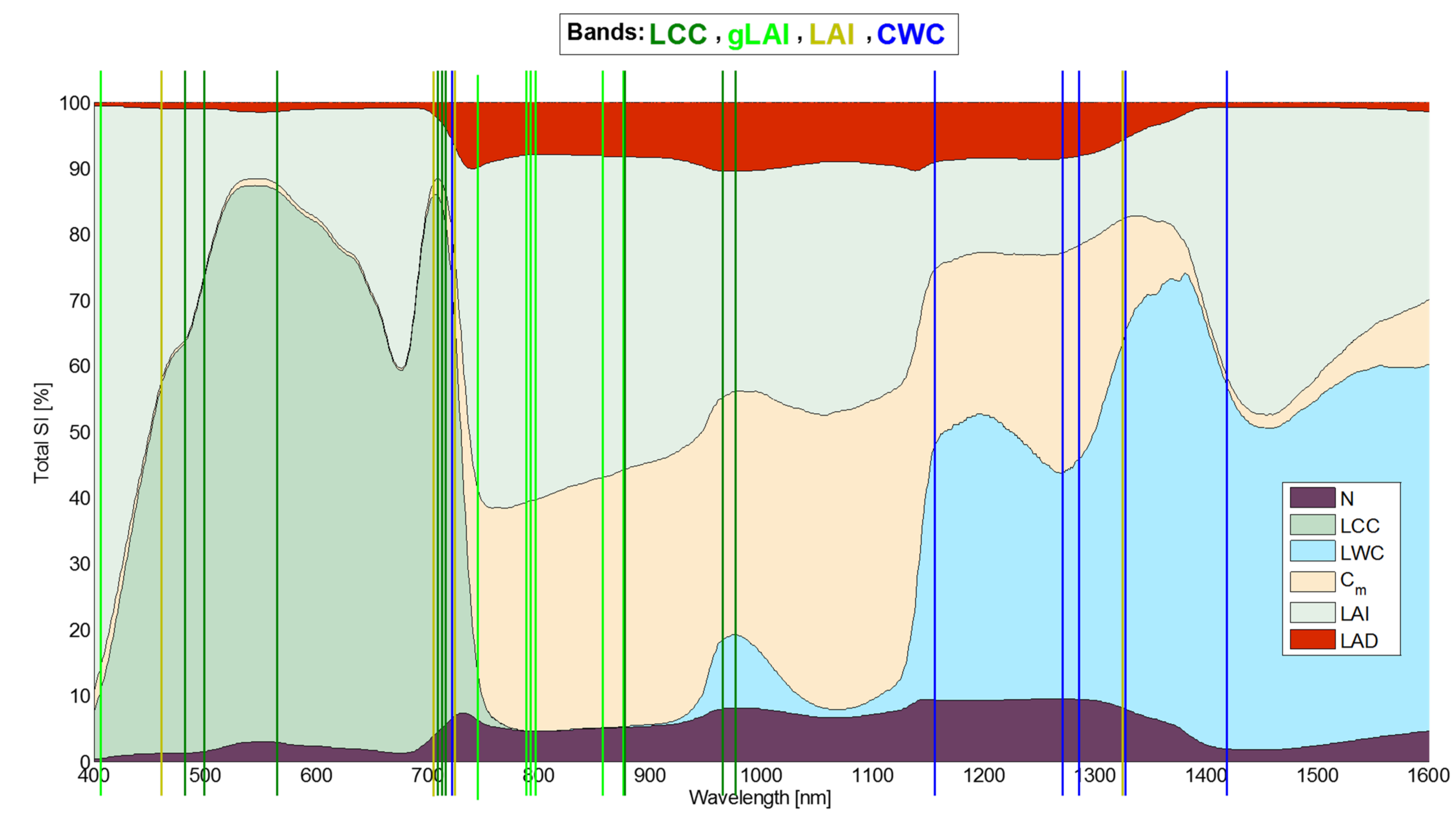} 
\end{center}
\caption{PROSAIL $S_{Ti}$ results [\%] for vegetation variables (only shown between 400-1600 nm for clarity). Added on top are the wavelengths as identified by top performing regression models for LCC (dark green) gLAI (bright green) LAI (brownish green) and CWC (blue) variables (see Tables \ref{USB2000_GPR_wvl} and \ref{HyMap_GPR_wvl}).}
\label{PROSAIL_GSA}
\end{figure}

To end with, the identified most sensitive wavebands for the best performing models (see Tables \ref{USB2000_GPR_wvl}, \ref{HyMap_GPR_wvl}) are interpreted against $S_{Ti}$ GSA results of the PROSAIL model. The $S_{Ti}$ of PROSAIL input vegetation variables along the 400-1600 nm spectral range is plotted in Figure \ref{PROSAIL_GSA}. It can be observed that this spectral range is mostly influenced by LCC in the visible (400-700 nm), LAI throughout the whole spectral range and dry matter ($C_m$) and LWC from the 750 nm and 950 nm onwards. Added on top are the GPR-identified sensitive wavelengths. Interpretation can be as follows:

\begin{itemize}
	\item Regarding LCC (field hyperspectra dataset; dark green lines), not surprisingly most sensitive wavebands are located within the dominant chlorophyll 400-750 nm absorption region. Three bands are found at the slope of the first peak and two bands at the second peak that ends in the red edge. Perhaps more surprising are the sensitive bands at the 878-980 nm region. This can be explained by mechanisms of variables co-variation. Measured leaf and canopy variables of the same target hold some dependency as vegetation variables are interrelated. Given that  both biochemical, biophysical and soil properties govern the TOC reflectance, co-variation relationships can play an important role in understanding spectral responses~\citep{Ollinger2011}. Regarding the LCC-related sensitive bands, note that at 878 nm $C_m$ and LAI are dominating, while at 966 and 980 nm there is also influence of LWC. Accordingly, spectral variability due to these variables that covary with LCC can support the estimation of LCC.
	\item Regarding gLAI (field hyperspectral dataset; bright green lines), one wavelength is at 406 nm and the others are in the red edge and in the 792-878 nm region. It can be observed that these are regions where the influence of LAI is maximized. In fact, it should be noted that the four sensitive bands in the 792-878 nm are positioned in a spectral region with very little variability. The difference in information content are minimal (if any). When resampling to a broader spectral bands will then essentially boil down to a band in the blue (if not affected by atmospheric effects), red edge and a band in the NIR, as was found most sensitive by~\citep{Kira2016}. Related to it, the PROSAIL GSA results also demonstrate why it is easier to predict LAI than other vegetation variables: LAI drives the spectral variation across the whole spectral range, with at various regions being the most dominant driver.
	\item Regarding LAI (HyMap dataset; brownish green lines), only four bands were needed to optimize LAI prediction: one at 462 nm that is driven mainly by LCC and LAI, two within the red edge (708, 723 nm) driven by LCC, LAI and $C_m$, and one at 1327 nm. The latter band is strongly affected by LCC, $C_m$ and LAI. Here again, given that the majority of the field samples were collected on irrigated parcels with green vegetation, covariance relationships play a role~\citep{Verrelst2012}.
	\item Regarding CWC (HyMap dataset; blue lines),	this is the product of LAI and LCC. In this respect, most of the sensitive bands are located within 1157-1419 nm where LCC has strong absorption regions and governs TOC reflectance. The band in the red edge (723 nm) can be explained by the relationship with LAI. Such covariations have also been demonstrated before~\citep{Ceccato2001,Zarco-Tejada2003,Yi2014}.
\end{itemize}

\section{Discussion}\label{sec:discussion}

By applying GPR-BAT to two different hyperspectral datasets, we have identified the most sensitive spectral bands in remotely predicting LCC and gLAI, LAI and CWC. The following general findings can be discussed:

\begin{itemize}
\item For both field and airborne hyperspectral datasets, using all spectral bands led to suboptimal regression performances. In case of narrowband field dataset (2 nm) it even performed almost as poor as when using only one band. Generally, entering all hyperspectral bands into a regression model is never a good idea due to multi-collinearity and inclusion of noisy bands. To overcome this, either band selection or spectral reduction techniques are recommended. On the other hand, as can be observed in the PROSAIL GSA results (Figure \ref{PROSAIL_GSA}), various vegetation properties are actually related to rather broad spectral regions (e.g. LAI, LCC, CWC). For mapping those variables resampling narrowband data to broader bandwidths can be beneficial, e.g. to \red{10 nm  as demonstrated by ~\citet{Kira2016}, or even to} the resolution of HyMap (bandwidths between 11 and 21 nm). Effectively, obtained HyMap results suggest that optimized LAI can be reached with few (i.e. 4) well-identified bands. 
   \item \red{Another aspect to be considered when analyzing relevant bands is the sginal-to-noise ratio (SNR). Most VNIR spectroradiometers employ silicon photodiode detectors that are characterized by a decreasing SNR at the shortest (i.e., $<$ 450 nm) and towards longer wavelengths (i.e., $<$ 900 nm))~\citep{milton2009}. That may explain the irregular $\sigma_b$ behavior from 900 nm onwards as observed in Figure \ref{sigmas_all}. In part, the lower $\sigma_b$ seem to be rather due to a poorer SNR than to information content related to vegetation properties. Therefore, instead of relying on single $\sigma_b$ results for band selection, a more robust method, e.g. such as GPR-BAT, is recommended.} 
   \item \red{GPR-BAT results suggest that band redundancy is less an issue when reducing to superspectral data (typically $<$ 50 bands), and using all those bands can equally lead to top-performing regression models. Accordingly, for multispectral and superspectral sensors of importance is to have the spectral bands rightly located along the spectral range (e.g., see assessment  of waveband performance for vegetation analysis applications by~\citet{Thenkabail2004}).} 
 \item For none of the tested variables using only two best-performing bands led to optimized results. This suggests that applying two-band indices to hyperspectral data is suboptimal and thus not recommended. \red{Earlier systematic vegetation indices studies confirm this observation~\citep{verrelst2015,Kira2016}, however they never went beyond three-band indices.} A full-spectrum band analysis, e.g. as the one proposed here, can identify more powerful band combinations.  
  \item 	For each of the variables, when considering the most sensitive bands, a band in the red edge region (700-750 nm) was found to be crucial. In this region both chlorophyll absorbance and structural variable govern TOC reflectance. The importance of the red edge region has been emphasized before~\citep{gitelson2003b,dash2004,Delegido11,Verrelst2012,clevers2012,Delegido2013,clevers2013,verrelst2015}.
   \item  Comparison against PROSAIL GSA results demonstrated that most of the identified best-performing bands are located within the variables dominant absorption regions. This suggests that the relationships identified by GPR-BAT have a physical meaning, as was also earlier observed in~\citep{Verrelst12a,Verrelst2012,Wittenberghe2014}. However, some relevant bands were located outside the expected absorption regions. It underlines the importance of variables co-variations that contribute to the strength of the regression models.  
\end{itemize}

\red{Overall, GPR-BAT can become valuable in rapidly identifying relevant absorption regions, as well relevant band combination for practical applications.
 Such automatically generated information is of interest in many applications, since using less, well-identified bands not only improve predictive accuracies but also increases processing speed especially when applied to hyperspectral data. For instance, analysis of the SPARC HyMap dataset did not take more than a few minutes, and the mapping itself took less than a minute. At the same time, identifying most sensitive bands is of interest in view of configuring optical sensors equipped with only a few bands, e.g. for UAV applications (see~\citet{Aasen2015,VonBueren2015} for a discussion).}  
With more bands included in the sequential process, computational time will last longer. To speed up processing the following options were implemented: (1) to remove multiple bands in the first iteration, or (2) to remove sequentially multiple least contributing bands. Particularly the first option seems attractive in case of working with hyperspectral data, since the majority of bands do not contribute to an optimized regression model.

Beyond the here presented remote sensing vegetation products, it would be interesting to apply GPR-BAT to spectral-variable datasets that include more subtle plant properties, such as leaf nitrogen content, biomass and primary production among others.  At the same time, there is no reason to restrict spectral analysis to reflectance measurements only. With latest airborne imaging spectrometers such as HyPlant~\citep{Rascher2015} and with ESA's next Earth Explorer FLEX mission~\citep{Kraft2012}, it is now becoming possible to retrieve sun-induced chlorophyll fluorescence emission together with reflectance data. Only lately the full broadband fluorescence signal is starting to be explored at the canopy scale (e.g.~\citep{Verrelst2015GSA}). GPR band analysis at the leaf scale revealed that specific spectral regions of the fluorescence signal contain relevant information related to biochemical properties~\citep{Wittenberghe2014}. Similarly, it is expected that GPR band analysis at the canopy scale will reveal and confirm relationships between specific fluorescence spectral regions and plant stress~\citep{Ac2015}.

\section{Conclusions}\label{sec:conclude}

With the  {purpose of identifying an optimized number of spectral bands for vegetation properties estimation from hyperspectral data, in this study we presented the implementation of Gaussian processes regression (GPR) band analysis tool (GPR-BAT) within ARTMO's MLRA toolbox. 
GPR-BAT sequentially removes the least contributing band during the development of a GP regression model until only one band is kept.}  
Goodness-of-fit validation and band ranking results are tracked and stored within a MySQL database. This procedure has been automated and made user-friendly in a GUI framework \red{and} enables the user to: (1) identify the most informative bands for a given surface geophysical or biophysical variable, and (2) find the least number of bands that preserve high predictive accuracy using a GPR model.

GPR-BAT was applied to two hyperspectral datasets: (1) a field hyperspectral 400-1000 nm dataset and associated field data of LCC and gLAI collected in two contrasting crops as for maize and soybean, and (2) an airborne HyMaP 430-2490 nm dataset with LAI and CWC collected on an agricultural test site covering various crop types. Results showed that using all hyperspectral wavebands did not lead to most accurate GP regression model. Especially when many bands are involved as in case of the field hyperspectral dataset (301 bands), reducing bands using GPR-BAT improved retrievals with a $R^2_{CV}$ from 0.55 to 0.79 for LCC and 0.88 to 0.94 for gLAI. Hence, selecting the most informative bands becomes strictly necessary.
In fact, for each of the considered variables top performances were found between four and nine bands, and all of them relied on a band in the red edge and other bands in relevant absorption regions. After having identified a best performing regression model, it can be applied to hyperspectral images for optimized and automated vegetation property mapping.

\section{Acknowledgements}\label{sec:acks}

This work was partially supported by the European Space Agency under project 'FLEX-Bridge Study' (ESA contract RFP IPL-PEO/FF/lf/14.687), and by the the European Research Council (ERC) under the ERC-CoG-2014 SEDAL grant. AG is thankful to Marie Curie International Incoming Fellowship for supporting this work. We are very thankful to the Center for Advanced Land Management Information Technologies (CALMIT) and the Carbon Sequestration Program, University of Nebraska-Lincoln for sharing the data. \red{We thank the two reviewers for their valuable suggestions.}




\end{document}